\documentclass[runningheads]{llncs}
\usepackage{graphicx}

\usepackage{tikz}
\usepackage{comment}
\usepackage{amsmath,amssymb} 
\usepackage{color}
\usepackage{xcolor}
\usepackage{algpseudocode}
\usepackage{xspace}
\usepackage{tabularx}
\usepackage[inline]{enumitem}
\usepackage{hyperref}

\usepackage{scalefnt}
\newcommand{\textmc}[1]{\textsc{\scalefont{0.9}#1}}

\newcommand*{\dataset}{\textmc{NewsStories}\xspace}
\newcommand{\articleset}{\mathcal{A}}
\newcommand*{\STDDEV}{Std.~dev.~}

\newcommand*{\jpeditok}[2]{#1\xspace} 

\newcommand*{\thomasedit}[2]{\textcolor{black}{#1\xspace}} %

\usepackage[accsupp]{axessibility}  

\makeatletter
\DeclareRobustCommand\onedot{\futurelet\@let@token\@onedot}
\def\@onedot{\ifx\@let@token.\else.\null\fi\xspace}
\def\eg{\emph{e.g}\onedot}

\newcommand{\imageset}{\mathcal{I}}
\newcommand{\loss}{\mathcal{L}}

\usepackage{xcolor}
\usepackage{xspace}
{}

\begin{document}
\pagestyle{headings}
\mainmatter
\def\ECCVSubNumber{6141}  

\title{{\LARGE \dataset}: \\ Illustrating articles with visual summaries} 

%

\titlerunning{NewsStories: Illustrating articles with visual
summaries}
\author{Reuben Tan$^{1}$\thanks{Work done as part of an internship at Google} \ \ \ \ Bryan A. Plummer$^{1}$ \ \ \ \ Kate Saenko$^{1}$\thanks{Also affliated with MIT-IBM Watson AI Lab} \ \ \ \ JP Lewis$^{2}$ \ \ \ \ Avneesh Sud$^{2}$ \ \ \ \ Thomas Leung$^{2}$ \\
$^{1}$Boston University, $^{2}$Google Research \\
{\tt \small \{rxtan, bplum, saenko\}@bu.edu}, {\tt \small \{jplewis, asud, leungt\}@google.com} \\
} 
\authorrunning{R.~Tan et al.}
\institute{}

\maketitle

\begin{abstract}
Recent self-supervised approaches have used large-scale image-text datasets to learn powerful representations that transfer to many tasks without finetuning. These methods often assume that there is a one-to-one correspondence between  images and their (short) captions. However, many tasks require reasoning about multiple images paired with a long text narrative, such as photos in a news article.
In this work, we explore a novel setting where the goal is to learn a self-supervised visual-language representation
from longer text paired with a set of photos, which we call \textit{visual summaries}.
In addition, unlike prior work which assumed captions have a \textbf{literal} relation to the image, we assume images only contain loose  \textbf{illustrative} correspondence with the text.  To explore this problem, we introduce a large-scale multimodal dataset called \dataset containing over 31M articles, 22M images and 1M videos. We  show that state-of-the-art image-text alignment methods are not robust to longer narratives paired with multiple images, and 
introduce an intuitive baseline that outperforms these methods, e.g., by 10\% on on zero-shot image-set retrieval in the GoodNews dataset\footnote{Project page: \url{https://github.com/NewsStoriesData/newsstories.github.io}}.

\end{abstract}

\section{Introduction}
\begin{figure*}[t]
    \centering
    \includegraphics[width=0.49\linewidth]{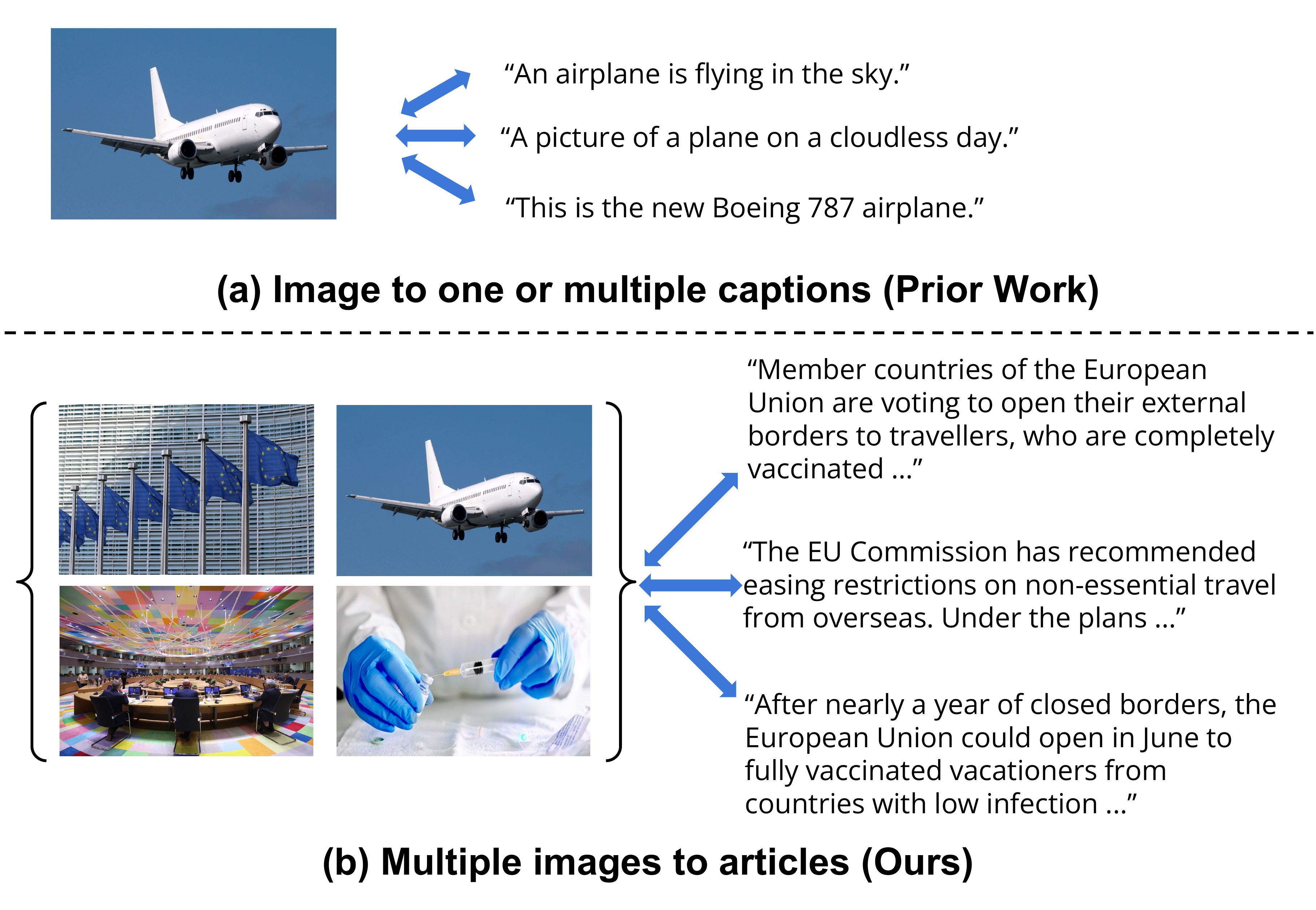}
    \includegraphics[width=0.47\linewidth]{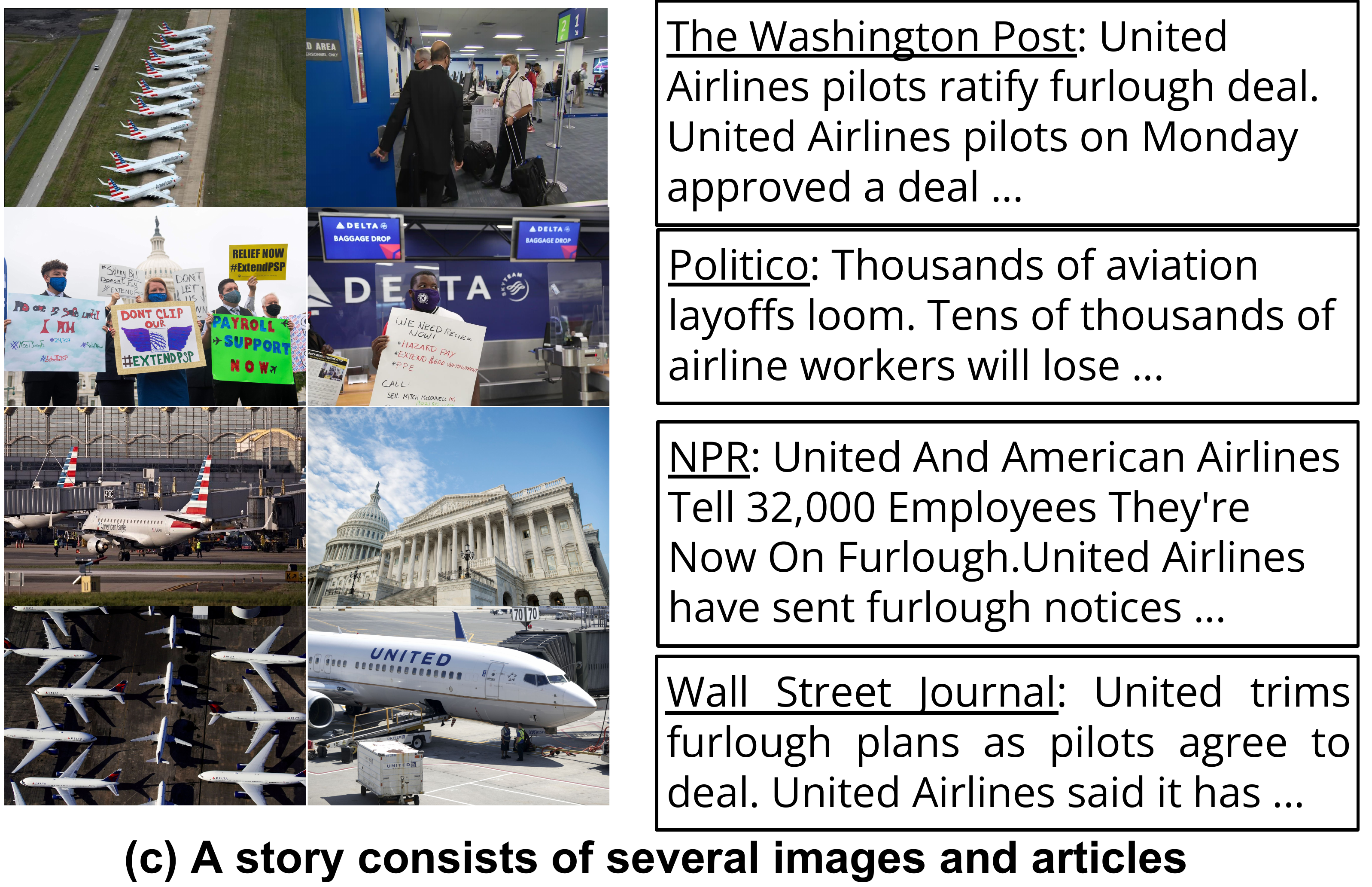}
    \vspace{-4mm}
    \caption{\small Unlike prior work (a) which aligns a single image with one or more captions, we study the problem of learning the multiplicity of correspondences between an unordered set of visually diverse images and longer text sequences (b). (c)~shows an example story from our \dataset dataset. For each  story, we cluster articles from different media channels and collect images that are used in the articles. In contrast to conventional \textbf{literal} caption datasets such as MSCOCO and Flickr30K, the images and text narratives in \dataset only have \textbf{illustrative} relationships}\vspace{0mm}
    \label{fig:motiv}
\end{figure*} 

State-of-the-art image-and-text representation learning approaches generally focus on learning a one-to-one correspondence between an image and one \cite{lee2018stacked,CLIP,ALIGN} or more captions \cite{PCME,PVSE,thomas2020preserving}, such as a photo with a caption \textit{``An airplane is flying in the sky"} (Figure~\ref{fig:motiv}a). While existing datasets such as MSCOCO \cite{lin2014microsoft} and Flickr30K \cite{young2014image} contain multiple captions for an image, the aforementioned approaches still learn a one-to-one correspondence between an image and a short caption that generally has a strong literal relation to it. However, this is unrealistic for longer text narratives containing multiple images (\eg news articles, Wikipedia pages, blogs). To challenge such constraints, Kim \emph{et al.} \cite{kim2015ranking} first introduce the problem of retrieving image sequences based on blog posts that may be composed of multiple paragraphs, using the assumption that the image sequence and the paragraphs have the same weak temporal ordering. However, this assumption is quite restrictive due to the prevalence of long narrative texts and groups of relevant images without information about their temporal order, \eg, in the news domain and Wikipedia. More importantly, they do not consider semantically related text narratives that may use similar groups of images.

Motivated by recent representation learning approaches which leverage large-scale data \cite{CLIP,ALIGN}, we seek to address the important problem of learning visual-semantic representations from text narratives of varying lengths and groups of complementary images from publicly available data. In this paper, we address this research problem in the news domain due to the prevalence of related articles and their corresponding images on the same story from different media channels. However, this is a general problem inherent in other domains including Wikipedia and social media posts on similar events. We define a \textit{story} as an event associated with a \textit{visual summary} consisting of images that illustrate it and articles that describe it, or videos depicting it. In contrast to prior work, this problem requires the capability to reason about the multiplicity of correspondences between sets of complementary images and text sequences of varying lengths. For example, in (Figure~\ref{fig:motiv}b), we aim to identify the story that is jointly illustrated by images of an airplane, flags of the European Union (EU) and a nurse preparing a vaccine. Here, one possibility is about traveling to the EU during a pandemic. A story could contain a variety of photos that illustrate a particular concept, and conversely, stock images of an airplane or EU flags could illustrate different stories.

We formulate the problem of visual summarization as a retrieval task, where the image sets are given and the goal is to retrieve the most \emph{relevant} and \emph{illustrative} image set for an article. Our proposed research problem is distinguished from prior work in two ways. First, we must be able to reason about the \emph{many-to-many} correspondences between multiple images and linguistically diverse text narratives in a story. Second, the images in our problem setting often only have \emph{illustrative} correspondences with the text rather than \emph{literal} connections (\eg ``travel'' or ``vacation'' rather than ``airplane flying''). Extracting complementary information from images and relating them to the concepts embodied by the story is a relatively under-explored problem, especially when the images and text only have loose and symbolic relationships. While existing work such as Visual Storytelling \cite{VisStorytell} aims to generate a coherent story given a set of images, the images have a temporal ordering and exhibit \emph{literal} relations to the text.

\begin{table}[t]
\centering
\caption{{\bf Dataset statistics comparison.} Our \dataset dataset is significantly larger than existing datasets with diverse media channels and story clusters that indicate related articles, images and news videos}
\vspace{-2mm}
 \begin{tabular}{| l | r | r | r | r | r |} 
 \hline
  & GoodNews & NYTimes & Visual News & NewsStories & NewsStories\\
  & \cite{GoodNews} & 800K~\cite{tran2020transform} & \cite{VisualNews} & Unfiltered & Filtered\\
 \hline
 \# Media channels & 1 & 1 & 4 & 28,215 & 46 \\
 \hline
 \# Story clusters & - & - & - & - & 350,000 \\
 \hline
 \# Articles & 257,033 & 444,914 & 623,364 & 31,362,735 & 931,679 \\
 \hline
 \# Images & 462,642 & 792,971 & 1,080,595 & 22,905,000 & 754,732 \\
 \hline
 \# Videos & 0 & 0 & 0 & 1,020,363 & 333,357 \\
 \hline
 Avg article length & 451 & 974 & 773 & 446 & 584 \\
 \hline
\end{tabular}
\label{tab:dataset_comparisons}
\end{table}

To facilitate future work in this area, we introduce \textbf{\dataset}, a large-scale multimodal dataset (Fig.~\ref{fig:motiv}-c and Table~\ref{tab:dataset_comparisons}). It contains approximately $31M$ articles in English and $22M$ images from more than $28k$ news sources. Unlike existing datasets, \dataset contains data consisting of three modalities - natural language (articles), images and videos. More importantly, they are loosely grouped into \emph{stories}, providing a rich test-bed for understanding the relations between text sequences of varying lengths and visually diverse images and videos. With an expanding body of recent work on joint representation learning from the language, visual, and audio modalities, we hope that our \dataset dataset will pave the way for exploring more complex relations between multiple modalities.

Our primary goal is to learn robust visual-semantic representations that can generalize to text narratives of varying lengths and different number of complementary images from uncurated data. We benchmark the capabilities of state-of-the-art image-text alignment approaches to reason about such correspondences in order to understand the challenges of this novel task. Additionally, we compare them to an intuitive Multiple Instance Learning (MIL) approach that aims to maximize the mutual information between the images in a set and the sentences of a related articles. We pretrain these approaches on our \dataset dataset before transferring the learnt representations to the downstream task of article-to-image-set retrieval on the GoodNews dataset under 3 challenging settings, without further finetuning. Importantly, we empirically demonstrate the utility of our dataset by showing that training on it improves significantly on CLIP and increases the robustness of its learnt representations to text with different numbers of images. To summarize, our contributions are as follows:
\begin{enumerate}
    \item We propose the novel and challenging problem of aligning a story and a set of illustrative images \emph{without temporal ordering}, 
    as an important step towards advancing general vision-and-language reasoning, and with applications such as automated story illustration and bidirectional multimodal retrieval.
    \item We introduce a large-scale news dataset \dataset that contains over 31M articles from 28K media channels as well as data of three modalities. The news stories provide labels of relevance between articles and images.
    \item We experimentally demonstrate that existing approaches for aligning images and text are ineffective for this task and introduce an intuitive MIL approach that outperforms state-of-the art methods as a basis for comparisons. Finally, we show that training on the \dataset dataset significantly improves the model's capability to transfer its learned knowledge in zero-shot settings.
\end{enumerate}
\section{Related Work}
To the best of our knowledge, there has been limited work that directly address our problem of learning  \emph{many-to-many} correspondence between images and texts. Wang et al. \cite{wang2016learning} propose to learn an alignment between image regions and its set of related captions. However, the images and captions in their setting have strong literal relationships instead of illustrative correspondences. Current vision-language models have other applications including text-based image retrieval \cite{young2014image},
visual question answering \cite{antol-vqa-15,zhu-visual7w-16} and visual reasoning \cite{suhr-visualreasoningcorpus-19},  and as a tool for detection of anomalous image-text pairing in misinformation \cite{aneja-misinfo,tanDIDAN2020}. 

Recent vision-language models \cite{CLIP,ALIGN} demonstrate excellent zero-shot performance on various downstream tasks,
sometimes exceeding the performance of bespoke models for these tasks.
This advancement has relied on very large-scale datasets consisting simply of images and their associated captions. Such datasets require little or no curation, whereas the need for training labels has limited the size of datasets in the past \cite{ConceptualCaptions}.
These image-caption datasets are paired with a natural contrastive learning objective, 
that of associating images with their correct captions \cite{ConVIRT,CLIP,ALIGN}.
Previous work has demonstrated that improved visual representations can be learned by predicting captions (or parts of them) from images \cite{joulin-visualfeatures-eccv16,li-ngrams-iccv17,sariyildiz-captions-eccv20,VirTex}.
Captions provide a semantically richer signal \cite{VirTex} than the restricted number of classes in a dataset such as ImageNet -- for example, a caption such as ``my dog caught the frisbee'' mentions two objects and an action. 

Closer to our work are methods that learn one-to-many correspondences from images or videos 
to captions \cite{PCME,PVSE,thomas2020preserving}, or vice-versa.  
Polysemous  Instance  Embedding  Networks (PVSE) \cite{PVSE}
represents a datum from either modality with multiple embeddings representing different aspects of that instance,
resulting in $n \times n$ possible matches. They use multiple instance learning (MIL) \cite{dietterich-MIL-97} align a image-sentence pair in a joint visual-semantic embedding \cite{frome-visualsemanticembedding-13,karpathy-visualsemanticalign-15,faghri-VSEplusplus-18} while ignoring mismatching pairs.
PCME \cite{PCME} explicitly generalizes the MIL formulation from a single best match to represent the set of possible mebeddings as a normal distribution, and optimize match probabilities using a soft contrastive loss \cite{oh-hib-2018}. 
In contrast to prior work, in our setting both the visual and text modalities contain semantically distinct concepts that are not appropriately represented with a unimodal probability density.

Finally, while there exist news datasets including GoodNews \cite{GoodNews}, NYT800k \cite{tran2020transform} and VisualNews \cite{VisualNews}, 
they are only sourced from a single (GoodNews and NYT800K) or four (VisualNews) media channels. 
Additionally, the VMSMO dataset \cite{li2020vmsmo} contains news articles and videos that are downloaded from Weibo, but it does not contain images. Compared to these datasets, our \dataset dataset not only contains articles from over 28K sources, but also has story labels to indicate related articles and images. Related to our work, \cite{headline2020} released the NewSHead dataset for the task of News Story Headline generation, containing $369k$ stories and $932k$ articles but no images. However, ours contains a much larger corpus of stories and associated articles. Last but not least, the aforementioned datasets generally only contain either images and articles or videos and articles. In contrast, ours provides data from all 3 modalities.

\section{The \dataset Dataset}
\label{sec:dataset}

Our \dataset dataset comprises the following modalities: 
\begin{enumerate*}
    \item news articles and meta data including titles and dates
    \item images
    \item news videos and their corresponding audio.\thomasedit{}{ and transcribed narrations}
\end{enumerate*}
As mentioned above, \dataset has three main differences from existing datasets. First, it is significantly larger in scale and consists of data from a much wider variety of news media channels. Second, unlike a significant percentage of multimodal datasets, it contains three different modalities -- text, image and videos. Third, the text, images, and videos are grouped into stories. This provides story cluster labels that not only help to identify related articles but also create sets of multiple corresponding images for each story.

\vspace{-3mm}
\subsection{Data collection}
Learning the multiplicity of correspondences between groups of complementary images and related text narratives requires a dataset that contains multiple \emph{relevant but different} images that correspond to a given text sequence and vice versa. Curating a large-scale dataset with these characteristics is an extremely expensive and time-consuming process. In contrast, the news domain provides a rich source of data due to the proliferation of online multimedia information. We collected news articles and associated media links spanning the period between October 2018 and May 2021\footnote{CommonCrawl \cite{commoncrawl} can be used to fetch web articles.}. The articles from a filtered subset (Section~\ref{sec: dataset_filtering}) are grouped into \emph{stories} by performing agglomerative clustering on learned document representations \cite{liu2021newsembed}, similar to~\cite{headline2020}, which iteratively loads articles published in a recent time window and groups them based on content similarity. We merge clusters that can possibly share similar images via a second round of clustering based on named entities. Specifically, we begin by extracting the named entities in the articles using Spacy and their contextualized representations with a pretrained named-entity aware text encoder \cite{yamada2020luke}. 
To obtain a single vector for the story cluster, we perform average-pooling over all named entity representations across all articles. 
Finally, these representations are merged into a slightly smaller number of clusters. We extracted  video links, however only the text and images are used for alignment in Section~\ref{sec:approach}.

\vspace{-3mm}
\subsection{Dataset filtering} \label{sec: dataset_filtering}
We observed that there is a large amount of noise in the collected datasets due to the prevalence of smaller media channels. To address this, we removed 
articles and links that do not belong to a curated list of 46 news media channels\footnote{https://www.allsides.com/media-bias/media-bias-chart} selected by an independent organization rating media biases. The list contains major news sources including BBC, CNN, and Fox News.

\vspace{-3mm}
\paragraph{Curated evaluation set.}
Due to the sparsity of suitable datasets for this task, we curate a subset of the story clusters and use it as our evaluation set for the proxy task of retrieving image sets based on long narrative textual queries. Out of the 350,000 story clusters in the filtered story clusters, we randomly select 5000 clusters with at least 5 images. 
In the news domain it is common that articles on the same story may use similar photos. 
For example, different articles on the covid vaccinations may use the same image of a vaccination shot. To ensure that we can visually discriminate between two stories, we adopt a heuristic to ensure that the images are as diverse as possible. We begin by computing a set of detected entities for each image using the  Google Web Detection API \footnote{https://cloud.google.com/vision/docs/detecting-web}. 

To generate a visually diverse image set for a story, we compute all possible combinations of five images from the entire set of images present in the story and compute the intersection set of all detected entities over the images within a combination. Finally, we select the combination with the smallest intersection set as the ground-truth (GT) image set for a story. During training, each set of images is randomly sampled from all available images in the story cluster.

\subsection{Quality of story clusters}
To evaluate the quality of these story clusters, we use qualified human raters to judge three aspects of our \dataset dataset. Each data sample is rated by three humans and the rating with the most votes is selected as the final score. We provide the instructions for these evaluations in the supplementary.

\paragraph{Relatedness of articles in a story cluster.} For an approach to learn a meaningful alignment between groups of images and groups of related text, it is necessary that the articles in a story cluster are mostly relevant to each other. We randomly sample 100 stories and provide each rater with up to 10 articles from each story. A story is rated as of good quality if at least 80\% of the articles are related to each other. Out of 100 randomly selected story clusters, raters determine that 82\% of them are of good quality. Note that we do not try to eliminate all noise in the story clusters since we do not have ground-truth targets.
\vspace{-2mm}
\paragraph{Relevance of images in a story cluster.} To rate the semantic relevance of the images to the story, each rater is provided with a maximum of 10 articles and 20 images from each cluster. The image set is labeled as relevant to the articles if at least 80\% of the images are plausible in the context of the story. 76\% of the randomly selected sets are rated as relevant. Some possible sources of irrelevant images may come from links to other articles or advertisements.
\vspace{-3mm}
\paragraph{Ambiguity of ground-truth image sets in the evaluation set.}
A well-known problem of existing bidirectional sentence-to-image retrieval datasets is that some sentences can be relevant to other images in the dataset that are not in the ground truth, which results in inherent noise in the evaluation metrics.  
We use raters to determine if humans are able to discriminate between the ground-truth image set and others that can potentially be relevant. We randomly sample 150 stories and use the pretrained CLIP model to rank the image sets given the query article.  A  rater is provided with the GT image set as well as the top-5 sets retrieved by the CLIP model and is asked to select the most relevant image set. 86\% of GT sets are selected by at least 2 out of 3 raters as the most relevant, indicating that our annotations are of high quality.


\subsection{Data statistics}
We compare our \dataset dataset to existing news datasets such as GoodNews, NYT800K and VisualNews \cite{GoodNews,VisualNews} in Table~\ref{tab:dataset_comparisons}. We present statistics of both unfiltered and filtered sets. In contrast to existing datasets, the entire dataset contains articles from approximately 28K news media channels, which significantly increases its linguistic diversity. Additionally, it contains over 31M articles and 22M images. We compute the story clusters over articles from the filtered set due to computational constraints, but we release the entire dataset. The articles, images, and videos in the final filtered set are grouped into approximately 350K fine-grained story clusters.  
The number of articles per story cluster varies greatly across different clusters, ranging from a minimum of 1 to a maximum of about 44,000 (see suppl. for frequency histograms). We observe that story clusters that contain unusually high number of articles tend to come from the entertainment domain, e.g., reality television shows. These story clusters are removed from the final filtered set to eliminate noise.  Additionally, we observe that most story clusters  contain about 1 to 20 images. This is indicative of the challenges faced in obtaining sufficient data to study the novel problem of learning multiple correspondences between images and text. Finally, please refer to the supplementary material for details on the video statistics.
\section{Illustrating articles with visual summaries}
\label{sec:approach}
\vspace{-2mm}

\begin{figure*}[t]
\begin{center}
\includegraphics[width=0.85\linewidth]{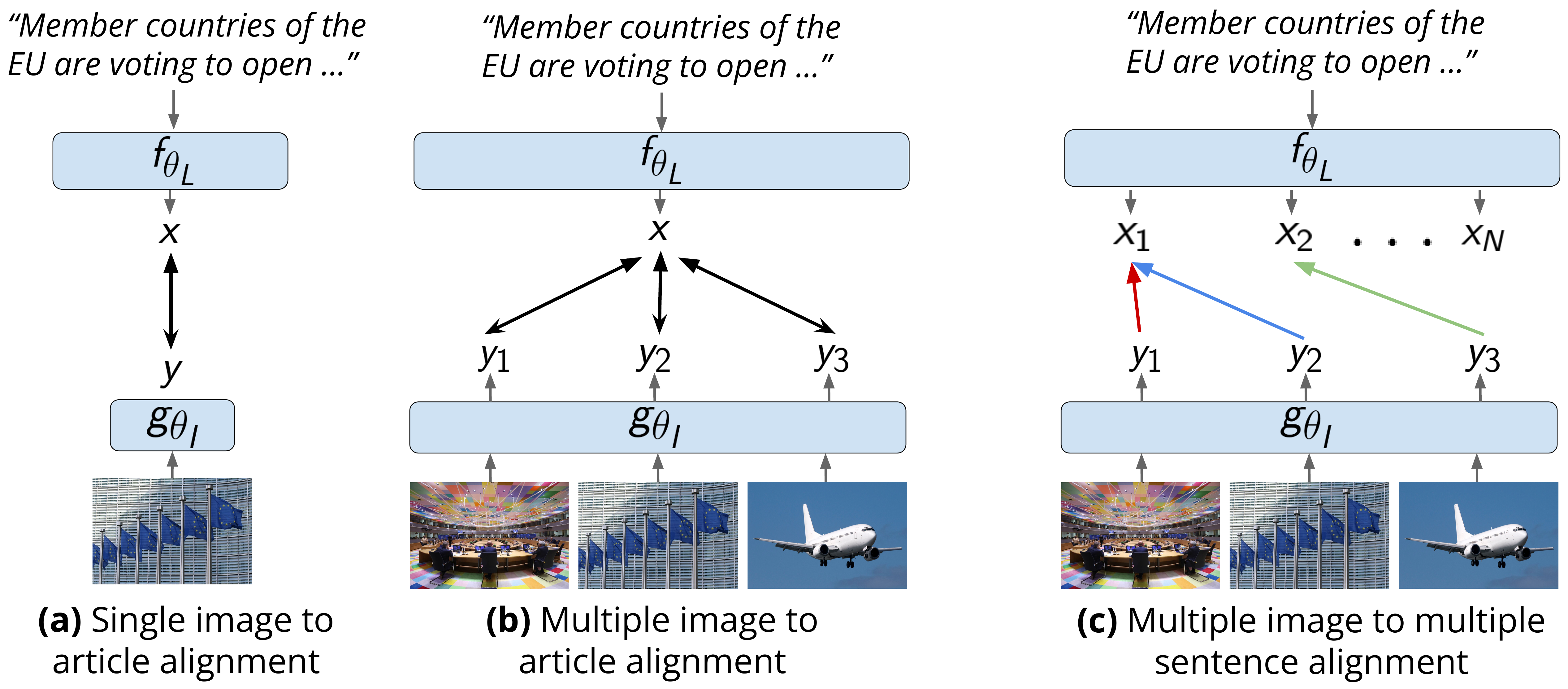}
\end{center}
\vspace{-6mm}
   \caption{ \textbf{Comparison of image-and-text alignment objectives.} The images and text are encoded by encoders  $g_{\theta_I}$ and $f_{\theta_L}$, respectively. 
   $x$ and $y$ denote the representations of the encoded article and image respectively. In (c), $x$ is also labeled with a subscript to indicate that it is the representation for a sentence in the article} 
\label{fig:models}\vspace{-0mm}
\end{figure*}

The primary objective of this work is to explore the problem of illustrating articles with visual summaries that comprise a varying number of images. By reasoning about the multiplicity of correspondences between longer text sequences and multiple image illustrations, we hope to jointly learn visual-semantic representations that are robust to text narratives of different lengths as well as a varying number of images.
Specifically, given a set of visually different images, the goal is to learn an aggregated and contextualized representation for the images such that it is able to infer the relevant story, regardless of the exposition styles in the articles.
To address this goal, we formulate the task of retrieving the most relevant set of images given a query article.

In this task, given a story  consisting of a set of related articles $\articleset$, and a corresponding set of images $\imageset$, where $ |\imageset| = N_I$, we aim to maximize the semantic similarity between each article $L \in \articleset$, and the entire image set $\imageset$. A language encoder $f_{\theta_L}$ is used to encode the entire text sequence $L$ into its representation $x \in \mathbb{R} ^ {N_L \text{ x } D}$. Depending on the number of text segments $N_L$, $L$ can be used to denote an article-level representation or a set of sentence representations.  Each image is encoded to obtain its base representation $y_i$ with an image encoder $g_{\theta_I}$ parameterized by weights $\theta_I$, where $i \in \{1, \cdot \cdot \cdot, N_I \}$.  

We describe several existing image-text alignment objectives that can be applied to our task in Sec.~\ref{sec:obj}. We then present a Multiple Instance Learning approach for maximizing the correspondence between an image and specific sentences in the article to determine the importance of fine-grained alignment for this problem in Sec.~\ref{milsim_obj}.
See Figure~\ref{fig:models} for a high level comparison of these objectives. An effective visual summarization of an article requires the set of retrieved images to be \emph{coherent} and \emph{complete}. Since negative image sets may only overlap with the story partially, we enforce coherence by constraining the models to rank them lower than the positive set for an article. Additionally, our proposed approach seeks to enforce completeness by maximizing
the semantic alignment between articles and the sets of relevant images via an article-level loss.

\vspace{-3mm}
\subsection{Existing alignment objectives}\label{sec:obj}

\paragraph{Single image-text contrastive loss:} We first explore an objective that aligns an image with a caption. Contrastive learning and its variants (InfoNCE and triplet loss) are commonly used to align an image with its corresponding text sequence by maximizing the similarity score between their respective representations~\cite{CLIP,ALIGN,FrozenInTime}. 
We  use the popular InfoNCE \cite{simclr,oord-cpcInfoNCE-18} loss, formulated as $\loss_{\text{InfoNCE}}(x, y)$:
\newcommand{\expt}[1]{\exp_{\tau}\!{#1}}
\newcommand*{\defeq}{\stackrel{\text{def}}{=}}     
\begin{equation}
  \begin{split}
     -\log & \frac{\exp{(sim(x, y)/\tau)}}{\exp{(sim(x, y)/\tau)} + \sum_{x'}\exp{(sim(x', y)/\tau)} + \sum_{y'}\exp{(sim(x, y')/\tau)}}
  \end{split}
  \label{infonce}
\end{equation}
where $x'$, $y'$  denote the non-corresponding text and image representations with respect to a ground-truth pair in a given batch, $\tau$ is a temperature value, and $sim(.)$ is a similarity function.

\paragraph{Multiple Instance Learning for Noise Contrastive Estimation:} The MIL-NCE formulation \cite{miech-MILNCE-cvpr20} provides a natural and intuitive extension to the regular InfoNCE objective by allowing for comparisons between a text sequence and multiple images, formulated as $\loss_{\text{MIL-NCE}}(x, y)$:
\vspace{-5pt}
 \begin{equation}
   -\log \frac{\sum_{i \in N_I}\exp{(sim(x, y_i) / \tau)}}{\exp{(sim(x, y) / \tau)} + \sum_{x'}\exp{(sim(x', y) / \tau)} + \sum_{y'}\exp{(sim(x, y') / \tau)}}
    \label{milnce}
 \end{equation}
When evaluating with sets of images for both the InfoNCE and MIL-NCE baselines, we first compute the similarity score between the text query and each image before taking their average as the final similarity score.

\paragraph{Soft contrastive loss:} PCME \cite{PCME} models each image or text instance as a probability distribution. The probability that an image and a text match ($m=1$) is computed as:
\begin{equation}
    \frac{1}{K^2}\sum_k^K \sum_{k'}^K p(m | z_I^k, z_L^{k'})
\end{equation}
where $z_I^k$ is the k-th sampled embedding from the image distribution and $K$ is the number of sampled embeddings from each modality. The probability that two sampled embeddings match is computed using:
$p(m | z_I^k, z_L^{k'} = \sigma(-\alpha \| z_I^k - x_L^{k'} \| + \beta)$, where $\sigma$ is the sigmoid function and $\alpha$ and $\beta$ are learnable parameters. Finally, the loss for a given pair of image $I$ and text $L$ is formulated as:
\begin{equation}
    \loss_{\text{soft}} = 
    \begin{cases}
    -\log p(m| I, L), & \text{if $I$ corresponds to $L$}\\
    -\log (1 - p(m| I, L)),              & \text{otherwise}
\end{cases}
\end{equation}
During training and evaluation, multiple representations are computed for each image and caption. The final similarity score between a \{image, caption\} pair is computed from the highest-scoring pair of image and caption representations.

\vspace{-3mm}
\subsection{Multiple Instance Learning - Sentence To Image (MIL-SIM)} \label{milsim_obj} \vspace{-1mm}
Inspired by \cite{kim2015ranking}, we assume that each image in a given set should correspond to at least one sentence in the the article. Consequently, we adopt a Multiple Instance Learning framework where a bag of image and sentence instances is labeled as positive if most of the instances are from the same story cluster and negative otherwise. The MIL-NCE loss formulation is a smooth approximation of the max function, thus it learns an alignment between the entire article and the most representative image in a set. In contrast, MIL-SIM tries to learn the illustrative correspondence between each image and the most relevant sentence. 

We segment the text article into individual sentences and encode them as $L=\{x_1, \cdots, x_{N_L}\}$.
Given a positive pair of image set $I = \{I_1, \cdot \cdot \cdot, I_{N_I}\}$ and text article $L$, we aim to maximize the mutual information between them:
\begin{equation}
    {\max\limits_{\theta} \mathbb{E} [\log \frac{p(I, L)}{p(I) p(L)}] },
\end{equation}
where $p(I)$, $P(L)$ and $P(I, L)$ are the marginal distributions for the images and articles as well as their joint distribution, respectively. 

In this setting, we do not have the ground-truth target labels which indicate the sentence that a given image should correspond to. This problem is compounded by the fact that some of the images may not originate from the same text source but from related articles. Consequently, we generate pseudo-targets by selecting the best matching sentence in an article for an image (colored arrows in Figure ~\ref{fig:models}(c)). Then we use Equation~\ref{infonce} to maximize the lower bound of the mutual information between them. Given an image representation $y_i$ and an article $L$, we compute their similarity as \jpeditok{$ \max_l x_l^T y_i$}{
\begin{equation}
    \max_l x_l^T y_i,
\end{equation}
}
where $l$ denotes the index of the sentence representation. In this formulation, multiple sentences in a corresponding article may be related to the image but will be treated erroneously as irrelevant. We circumvent this by selecting the highest-scoring sentences, with respect to the image, in articles from other clusters as negatives. Additionally, we mitigate the possibility of a different cluster containing a related sentence by reducing the weight of the image-sentence loss.

However, this may introduce a lot of noise to the learning process, especially if an image originates from a weakly-related article. To alleviate this problem, we impose an article-level loss that aims to maximize the general semantic similarity between the entire article and image set. We compute a single representation for the entire article $L_f$ as well as image set $I_f$ by mean-pooling over the sentence and image representations, respectively. Finally, we learn an alignment between them by minimizing the value of $\text{InfoNCE}(I^f, L^f)$, where their similarity is computed as: $\text{sim}(I^f, Y^f) = (I^f)^T Y^f$. Our final objective function is formulated as: 
\begin{equation}
    \loss_{\text{MIL-SIM}} = \sum_{b=1}^{B} L_{\text{InfoNCE}}(I_b^f, L_b^f) + \lambda * \sum_{b=1}^{B} \sum_{i=1}^{N_I} L_{\text{InfoNCE}}(I_{b,i}, L_b),
\end{equation}
where $B$ and $\lambda$ are the batch size and trade-off weight, respectively.

\section{Experiments}

\subsection{Implementation details}
We use the visual and text encoders of CLIP \cite{CLIP} as a starting point and finetune them using the image-and-text alignment objectives described above. The original CLIP model is trained end-to-end using 400 million image-and-text pairs. Due to the scale of the pretraining dataset, its representations have been demonstrated to be transferable to downstream tasks without further finetuning on the target datasets. During training, we finetune the projection layers of the CLIP encoders on the train split of our \dataset dataset. 
We extend the max input text length in CLIP from 77 to 256 in our experiments. We set an initial learning rate of 1e-5 and optimize the model using the Adam \cite{kingma-adam-15} optimizer, with a linear warm-up of 20,000 steps. The learning rate is gradually annealed using a cosine learning rate scheduler. We tune the hyperparameter settings by averaging the validation performance over 5 splits of 1000 articles that are randomly selected from the entire training set. In the MIL-SIM objective, we use the NLTK \cite{journals/corr/cs-CL-0205028} library to tokenize the articles into sentences and set the value of $\lambda$ to 0.1. 

\begin{table*}[t!]
\centering
\setlength{\tabcolsep}{4pt}
\caption{Comparison on the task of article-to-image-set retrieval on the test split of our \dataset dataset, which contains 5000 unseen stories with image sets of size 5. Higher R@K values and lower median rank indicate more accurate retrievals}
\vspace{-2mm}
  \addtolength{\tabcolsep}{1pt} 
  \begin{tabular}{|l|c|c|c|c|c|}
  \hline
    Method & Alignment & R@1 & R@5 & R@10 & Median\\
     & Type &  &  &  & Rank\\
    \hline
    Pretrained CLIP \cite{CLIP} & Single & 31.03 & 53.87 & 63.53 & 4 \\
    Single Image & Single & 35.88 & 63.58 & 74.12 & 3\\
    MIL-NCE \cite{miech-MILNCE-cvpr20} & Single & 32.84 & 59.60 & 70.92 & 3\\
    PVSE \cite{PVSE} & Single & 36.09 & 64.26 & 74.90 & 3\\
    PCME \cite{PCME} & Single & 35.18 & 65.52 & 75.65 & 3\\
    Transformer \cite{vaswani2017attention} & Multiple & 50.08 & 78.79 & 86.10 & 2\\
    Mean & Multiple & 49.12 & 76.04 & 85.18 & 2\\
    MIL-SIM & Multiple & \textbf{54.24} & \textbf{82.76} & \textbf{90.38} & \textbf{1}\\
    \hline
  \end{tabular}
  \label{tab:models_comparison}
\end{table*}

\begin{table*}[t]
\centering
\setlength{\tabcolsep}{1pt}
\caption{Zero-shot evaluations of article-to-image-set retrieval approaches on our test splits of the GoodNews \cite{GoodNews} dataset. Each test split has 3, 4, or 5 images in each article}
\vspace{-2mm}
  \begin{tabular}{|l|ccc|ccc|ccc|p{5mm}p{5mm}p{5mm}|}
  \hline
  & \multicolumn{3}{c|}{R@1} & \multicolumn{3}{c|}{R@5} & \multicolumn{3}{c|}{R@10} & \multicolumn{3}{c|}{Median Rank}\\
  \cline{2-13}
    Method  & 3 & 4 & 5 & 3 & 4 & 5 & 3 & 4 & 5 & 3 & 4 & 5\\
    \hline
    CLIP \cite{CLIP} & 22.29 & 21.13 & 20.90 & 41.14 & 39.25 & 38.83 & 49.94 & 47.33 & 47.41 & 11 & 13 & 13\\
    Single Image  & 17.27 & 16.27 & 15.61 & 34.84 & 32.57 & 32.55 & 43.94 & 41.47 & 40.95 & 16 & 19 & 20\\
    MIL-NCE \cite{miech-MILNCE-cvpr20}  & 13.75 & 13.87 & 13.30 & 30.14 & 29.06 & 28.33 & 38.30 & 37.40 & 36.40 & 24 & 26 & 29\\
    PVSE \cite{PVSE}  & 19.21 & 20.29 & 20.17 & 38.52 & 37.72 & 39.21 & 47.72 & 48.04 & 49.17 & 14 & 14 & 13 \\
    PCME \cite{PCME} & 20.08 & 20.65 & 20.14 & 39.36 & 39.72 & 39.91 & 48.12 & 48.56 & 49.03 & 14 & 14 & 13 \\
    Transformer \cite{vaswani2017attention}  & 29.06 & 28.69 & 29.41 & 51.15 & \textbf{50.77} & \textbf{51.61} & 59.83 & 59.71 & 60.57 & 5 & 5 & 5\\
    Mean &  28.73 & 28.01 & 28.77 & 50.22 & 49.11 & 50.24 & 58.80 & 58.64 & 59.39 & 5 & 6 & 5 \\
    MIL-SIM & \textbf{29.42} & \textbf{30.59} & \textbf{30.23} & \textbf{52.07} & 49.82 & 51.44 & \textbf{60.51} & \textbf{61.73} & \textbf{62.58} & \textbf{4} & \textbf{4} & \textbf{5}\\
    \hline
  \end{tabular}
  \label{tab:goodnews_comparisons}
\end{table*}

\begin{table*}[t]
\centering
\setlength{\tabcolsep}{2pt}
  \caption{Evaluation of retrieval models on article-to-image-set retrieval on the GoodNews dataset, where the candidate image sets do not contain a fixed number of images}
  \vspace{-2mm}
  \begin{tabular}{|l|c|c|c|c|c|}
  \hline
  Method & Alignment Type & R@1 & R@5 & R@10 & Median Rank\\
    \hline
    Pretrained CLIP \cite{CLIP} & Single & 18.43 & 36.59 & 46.92 & 12 \\
    Single Image \cite{CLIP} & Single & 17.14 & 33.77 & 43.56 & 17 \\
    MIL-NCE \cite{miech-MILNCE-cvpr20} & Single & 15.50 & 28.96 & 37.60 & 24 \\
    PVSE & Single & 18.57 & 37.70 & 47.78 & 12 \\
    PCME & Single & 19.37 & 39.19 & 48.04 & 12 \\
    Mean & Multiple & 21.30 & 41.56 & 51.66 & 9\\
    Transformer & Multiple & 20.30 & 40.88 & 49.24 & 11 \\
    MIL-SIM & Multiple & \textbf{25.12} & \textbf{46.17} & \textbf{56.16} & \textbf{7} \\
    \hline
  \end{tabular}
  \label{tab:third_eval_comparison}\vspace{0mm}
\end{table*}

\subsection{Evaluation Datasets and Metrics}
We conduct an evaluation of article-to-image set retrieval on both our proposed \dataset dataset and the GoodNews \cite{GoodNews} dataset, which contains approximately 250K articles from the New York Times. For our evaluation on the  GoodNews dataset, we create three different evaluation sets of 5000 articles with 3, 4 and 5 images with no overlapping articles or images between the sets and evaluate  on each separately. Note that compared to~\cite{thomas2020preserving} which retrieves the ground-truth image from a set of five images, our evaluation setup is more challenging and arguably more realistic for real-world applications.
We use two metrics: recall at top-K (R@1, R@5, R@10), where higher recall is better, and Median Rank of the correct sample, where lower is better.
\vspace{-20pt}
\subsection{Quantitative results} 
\noindent\textbf{\dataset.}
Table~\ref{tab:models_comparison} reports the Recall@K retrieval accuracies and median rank on the test split of our \dataset dataset. In this setting, each image set has a fixed size of 5 images. As demonstrated by Radford et al.\cite{CLIP}, the learnt representations of the pretrained CLIP model transfer effectively to the GoodNews dataset without further finetuning, obtaining a R@1 accuracy of 31.03\%. We observe that approaches that align a single image with a text sequence generally perform worse than variants that learn an aggregation function over the images. 

In contrast to the video variant of the MIL-NCE approach \cite{miech-MILNCE-cvpr20} which reports that aligning a video clip to multiple narrations leads to better performance on downstream tasks, applying such an approach on images and text that only have loose topical relationships does not work out-of-the-box. The retrieval accuracies obtained by MIL-NCE show that maximizing the similarity between the text representation and the most representative image in the set performs worse than training with single images. Despite using a simple average-pooling function, the mean images baseline outperforms the single image and MIL-NCE baselines significantly. This suggests that context between images is crucial. 

Although transformers have been shown to be effective at reasoning about context, using an image transformer to compute contextual information across the images only improves 1\% over the mean baseline. This indicates that the self-attention mechanism alone is insufficient to capture the complementary relations between related but visually different images. Last but not least, the significant improvements obtained by MIL-SIM over other alignment objectives highlight the importance of maximizing the alignment between an image and the most relevant segment in an article, despite not having access to the ground-truth pairings during training. We also provide results of an ablation study over the length of the input text sequence in the supplementary.

\begin{figure}[t!]
\begin{center}
\includegraphics[trim=0em 0em 0em 1em, clip, width=0.92\linewidth, height=4.6cm]{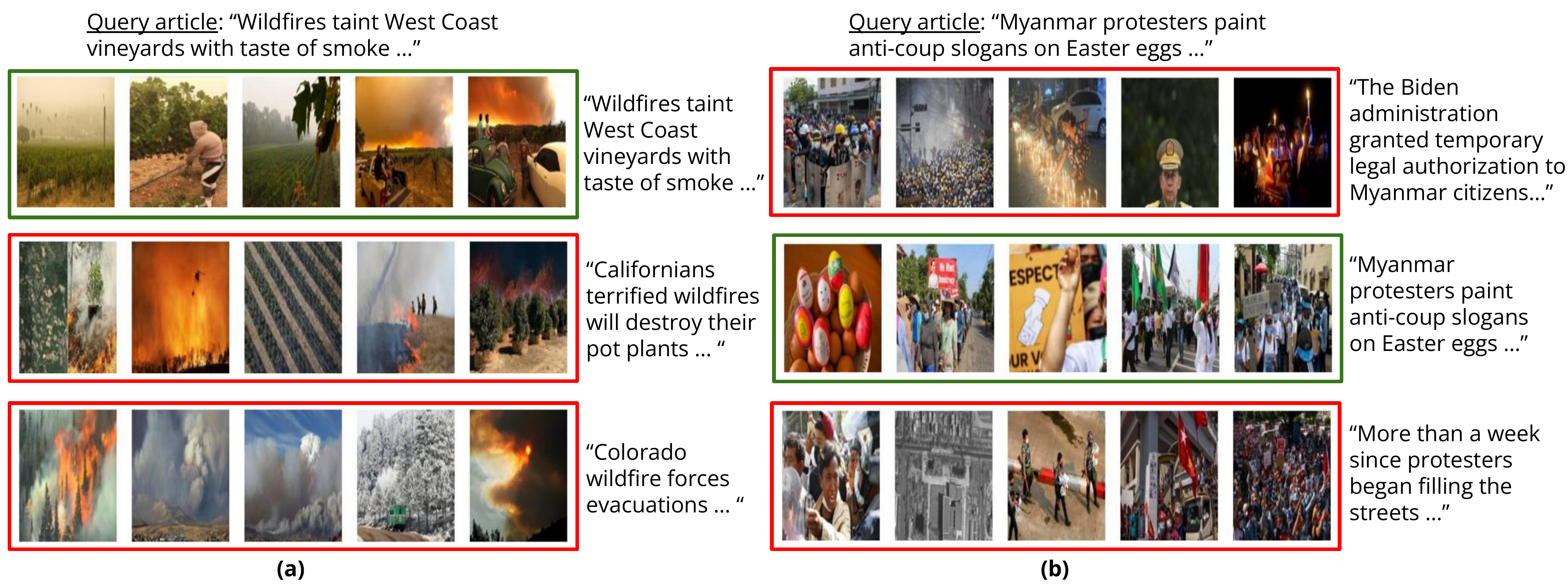}
\vspace{-8mm}
\end{center}
   \caption{Qualitative results showing the three top-ranked image sets per query on the test split of our \dataset dataset. The ground-truth and incorrect image sets, as determined by the cluster labels, are outlined with green and red boxes, respectively} \vspace{-8mm}
\label{fig:qualitative}
\end{figure}

\textbf{Zero-shot evaluation on GoodNews.} Next, we compare the effectiveness of our finetuned models on the curated test splits of the GoodNews dataset without further finetuning. We verify that none of the images and articles are present in our \dataset train split. Since GoodNews does not group articles into stories, the images in a set are obtained from the same article, instead of related articles as done in \dataset. Similar to the evaluation results on our \dataset dataset, Table~\ref{tab:goodnews_comparisons} shows that the pretrained CLIP model already performs well on all 3 test splits of the GoodNews dataset, achieving an average of approximately 21\% Recall@1 accuracy without any finetuning on the target dataset.  Additionally, finetuning the CLIP model on \dataset using the standard single image-and-text and MIL-NCE objectives actually leads to significant drops in performance from the pretrained CLIP model. These results suggest that learning a one-to-one correspondence is not optimal for allowing a model to reason about correspondences with multiple related images. This is corroborated by the observation that models trained to learn a one-to-one correspondences between images and text generally perform worse as the number of images increases. In contrast, we observe that training the models to align text with \emph{varying} numbers of images helps them to generalize to sets of images better.

By finetuning the CLIP model on our dataset, the mean baseline shows significantly improved retrieval accuracies despite not observing any articles and images from the GoodNews dataset during training. The much improved performances obtained by the mean images and transformer approaches demonstrate the importance of understanding the complementary relationships between images even in this setting, where all ground-truth images originate from the same text narrative. Similar to the results in the first setting, the results of the best-performing MIL-SIM approach suggests that being able to learn a mapping between each image and specific parts of the text narrative is crucial during training for this research problem, even if the images and text are only weakly-related.

\noindent\textbf{Zero-shot with multiple set sizes on GoodNews.} Finally, we evaluate on images sets of variable size in Table~\ref{tab:third_eval_comparison}, where the number of images in a set varies from 3 to 5. This requires a model to not only reason about the general semantic similarity between the query article and images but also determine if the number of images in a given set provides enough complementary information to discriminate one story from another. To this end, we randomly select 1500 articles from each of the above-mentioned GoodNews evaluation sets to create an evaluation split with different number of images per set.

Despite the inherent noise in obtaining positive text and image pairs using unsupervised clustering, the results suggest that aligning text narratives with varying numbers of complementary images during training is beneficial for learning more robust representations. These learnt representations are better able to discriminate between articles with different number of images.

\subsection{Qualitative results}
Figure~\ref{fig:qualitative} provides an example of correct and incorrect retrievals on the test split of our \dataset dataset. For each query article, the top 3 retrieved images are displayed in row order from top to bottom. Interestingly, in Figure~\ref{fig:qualitative}a, our mean image model is able to retrieve other set of images that are relevant to the notion of ``fire'', despite the fact that they belong to relatively different stories. Figure~\ref{fig:qualitative}b shows a hard example since the other two retrieved image sets are related to the query article, with the exception of not containing the image of the easter eggs. We include more retrieval visualizations in the supplementary. 

\subsection{Practical application of retrieving sets of images}
While we formulate the evaluation of this research problem as an article-to-image-set retrieval task, it may be impractical to find suitable images that have already been grouped into sets. Hence, we present an algorithm to find individual candidate images before grouping them into sets and ranking them using our trained models. We refer interested readers to the supplementary for more details as well as visualizations of the retrieval results.
\section{Conclusion}
In this work, we propose the important and challenging problem of illustrating stories with visual summarizes. This task entails learning many-to-many \emph{illustrative} correspondences between relevant images and text. To study this problem in detail, we introduce a large-scale multimodal news dataset that contains over 31M news articles and 22M images. Finally, we benchmark the effectiveness of state-of-the-art image-and-text alignment approaches at learning the many-to-many correspondences between the two modalities and draw useful insights from our empirical results for future research in this direction. 

\paragraph{Limitations and societal impact.}
Data and algorithms dealing with media can potentially be repurposed for misinformation. The selection of media channels in \dataset reflects the judgment of an independent organization. Our models are trained on top of the CLIP model and may inherit its bias (if any).

\noindent 
\textbf{Acknowledgements}: This material is based upon work supported, in part, by DARPA under agreement number HR00112020054.

%
%
\bibliographystyle{splncs04}
\bibliography{egbib}

\clearpage
\appendix
\def\fullpagefigurepad{8cm}


In this supplementary material, we provide the following:
\begin{enumerate}
    \item Additional statistics of the images, articles and videos included in our \dataset dataset.
    \item Ablation experiments over the length of the input text sequence using the best-performing MIL-SIM model.
    \item An algorithm for generating suitable image sets to describe any text narrative or article.
    \item Additional qualitative retrieval results on the test splits of the GoodNews dataset.
\end{enumerate}

To begin, we compute and present histograms over the number of articles, images and videos that are present in the computed story clusters in Section~\ref{newsstories_stats_supp}. Next, we conduct an ablation study in Section~\ref{text_ablation_supp} over the length of the input text sequences in the MIL-SIM approach to determine the importance of using additional context for learning the correspondences between text narratives and groups of complementary images. In Section~\ref{retrieval_algorithm_supp}, we describe an algorithm that an author may use to leverage the finetuned models to select a visually illustrative set of images for a given text narrative. Finally, we provide additional qualitative retrieval results of the MIL-SIM model on the test splits of the \dataset and GoodNews datasets in Section~\ref{retrieval_visualizations_supp}.

\section{Additional statistics of the NewsStories dataset} \label{newsstories_stats_supp}

\begin{figure}[h]
 \begin{center}
 \includegraphics[width=0.8\linewidth]{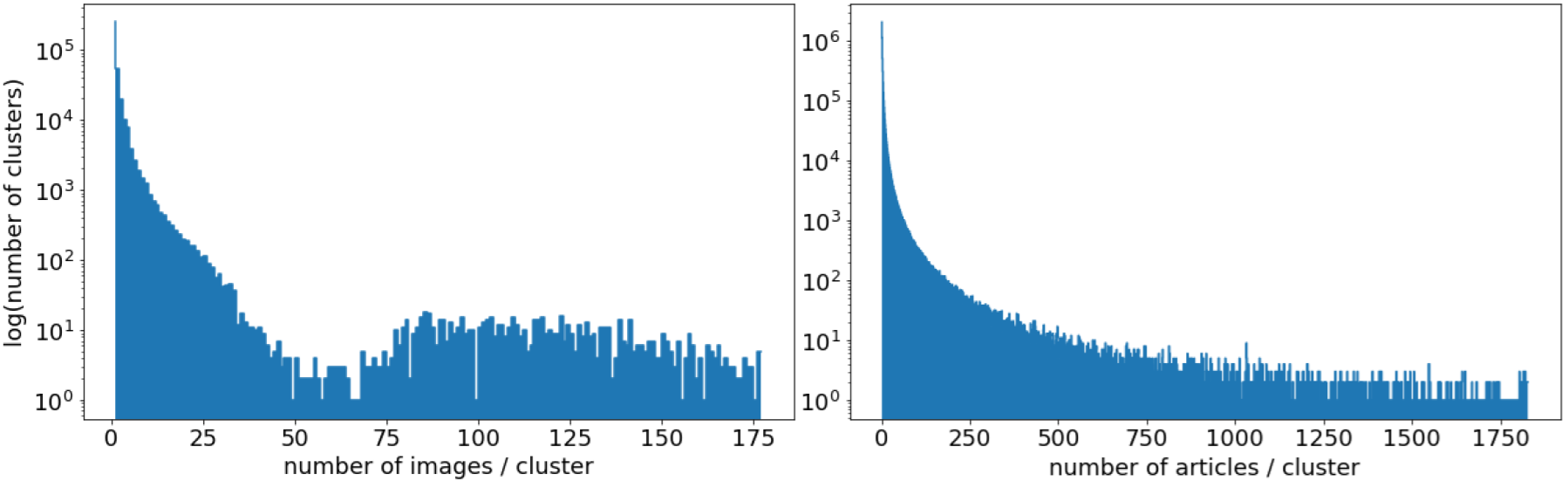}
 \end{center}
   \caption{Histogram of the number of images / articles that are contained in each news story cluster. For improved visualization, we show first $n$ bins that capture $95\%$ of total image / articles}
 \label{fig:statistics}
 \end{figure}

\paragraph{Article and images statistics. } Figure~\ref{fig:statistics} shows the histograms of the number of images and articles that are contained in each news story cluster. Note that we only show the first \emph{n} bins that contain 95\% of all images and articles for conciseness. As mentioned in the main paper, the number of articles per story cluster varies greatly across different clusters. Story clusters with unusually high numbers of articles are generally very noisy and tend to revolve around the theme of entertainment such as reality television updates or music videos. Additionally, a high percentage of story clusters contain between 1 and 20 images, where some of the images sourced from different media channels on the same story may be near-duplicates of each other. However, we note that such noise is prevalent in uncurated real-world data and being able to leverage these publicly available data to address the proposed research problem effectively remains an open question.

\begin{figure}
\centering
  \includegraphics[trim=40 20 0 0, clip,width=0.4\textwidth]{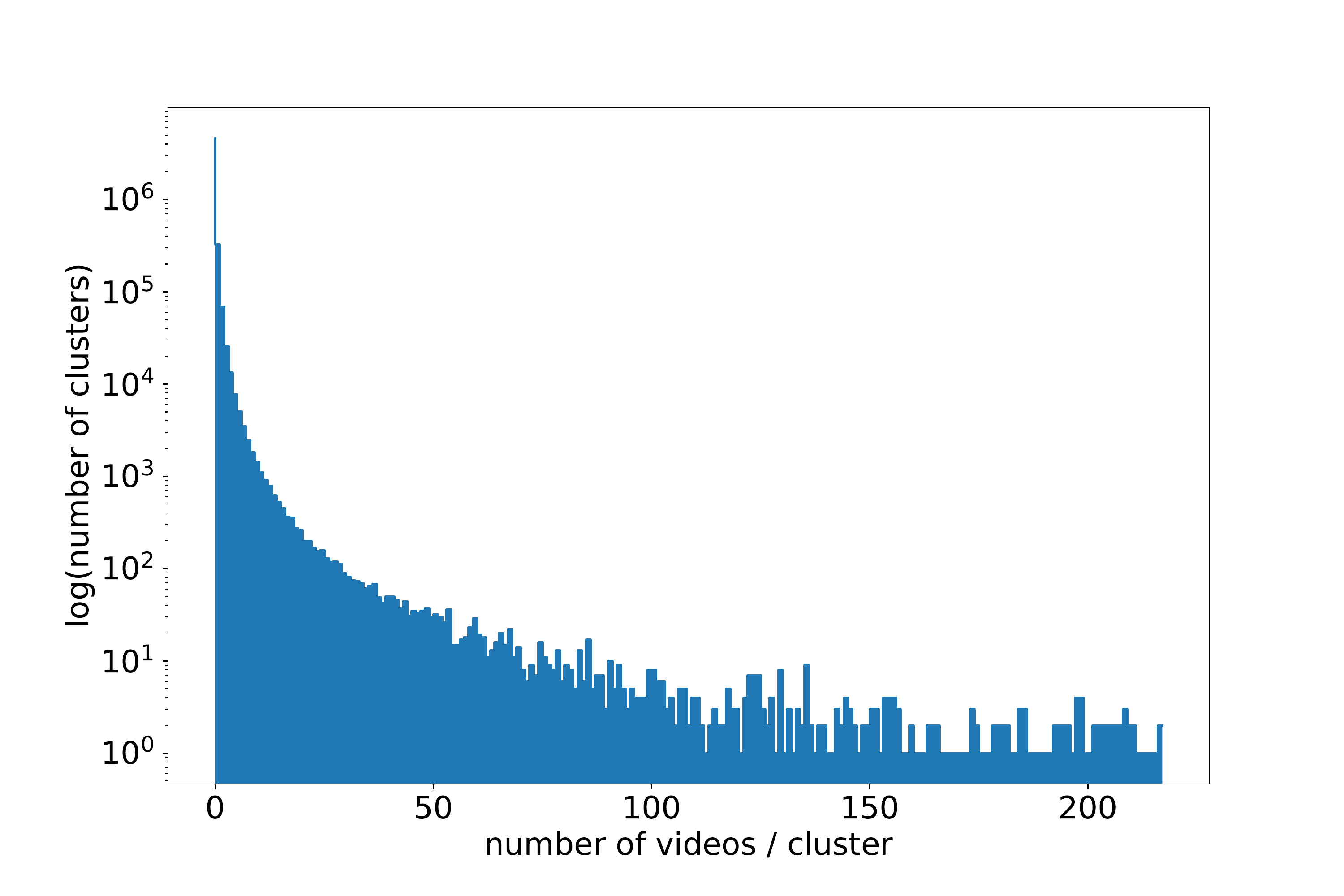}%
  \vspace{-1em}
  \caption{Histogram of number of videos per story cluster.}
  \label{fig:video_histogram}
\end{figure}

\paragraph{News videos statistics. } We provide a histogram of the number of videos contained in each story cluster of our unfiltered \dataset dataset in Figure~\ref{fig:video_histogram}. As corroborated by the data in Table~\ref{table:video_statistics}, we observe that a large percentage of news story clusters do not contain any videos at all. Additionally, the number of videos in a news story cluster varies considerably, from a minimum of zero to a maximum of 4005 videos. However, with over 450K stories containing at least one corresponding video, our \dataset dataset still provides a rich environment for learning to reason about multimodal correspondences between text, images, videos and audio.


\begin{table}
  \caption{Video story cluster statistics}
    \centering
\begin{tabular}{| l | r | r |} 

 \hline
 & Unfiltered & Filtered \\
 \hline
 Min.~\# videos in a cluster & 0 & 0\\
 \hline
 Max.~\# videos in a cluster & 4005 & 2255\\
 \hline
 \STDDEV \# videos per cluster & 4.22 & 6.79\\
 \hline
 Mode \# videos per cluster & 0 & 0\\
 \hline
 \# clusters with $\ge$ 1 video & 451,228 & 81,957\\
 \hline
\end{tabular}

  \label{table:video_statistics}

\end{table}

\begin{table*}[h!]
  \caption{Ablation of our best-performing MIL-SIM approach over the length of the input text sequences for the MIL-SIM approach on the validation split of our \dataset} 
  \begin{center}
  \begin{tabular}{|c|ccc|ccc|ccc|p{5mm}p{5mm}p{5mm}|}
  \hline
  Input Text & \multicolumn{3}{c|}{R@1} & \multicolumn{3}{c|}{R@5} & \multicolumn{3}{c|}{R@10} & \multicolumn{3}{l|}{Median Rank}\\
  \cline{2-13}
    Length & 3 & 4 & 5 & 3 & 4 & 5 & 3 & 4 & 5 & 3 & 4 & 5\\
    \hline
    32 & 34.28 & 36.46 & 38.08 & 56.66 & 60.19 & 62.64 & 65.11 & 68.52 & 70.43 & 3 & 3 & 3\\
    64 & \textbf{41.10} & 45.31 & 48.04 & 65.78 & 70.02 & 73.37 & 74.36 & 78.42 & 81.10 & 2 & 2 & 2\\
    128 & 39.96 & 45.48 & 48.58 & 67.01 & 72.15 & 75.74 & 76.47 & 80.65 & 84.02 & 2 & 2 & 2\\
    256 & 41.06 & \textbf{45.91} & \textbf{49.23} & \textbf{67.16} & \textbf{73.09} & \textbf{76.35} & \textbf{77.22} & \textbf{81.60} & \textbf{84.70} & 2 & 2 & 2\\
    512 & 39.44 & 44.88 & 48.73 & 66.39 & 72.29 & 75.74 & 76.50 & 81.44 & 84.69 & 2 & 2 & 2\\
    \hline
  \end{tabular}
  \end{center}
  \label{tab:text_len_ablation}
\end{table*}

\begin{table*}[t]
  \caption{Ablation of MIL-SIM over the length of input text sequences on zero-shot evaluations of article-to-image-set retrieval approaches on the GoodNews \cite{GoodNews} dataset. Each test split has 3, 4, or 5 images in each article.}
\setlength{\tabcolsep}{2pt}
  \begin{center}
  \begin{tabular}{|l|ccc|ccc|ccc|p{5mm}p{5mm}p{5mm}|}
  \hline
  Input Text & \multicolumn{3}{c|}{R@1} & \multicolumn{3}{c|}{R@5} & \multicolumn{3}{c|}{R@10} & \multicolumn{3}{c|}{Median Rank}\\
  \cline{2-13}
    Length  & 3 & 4 & 5 & 3 & 4 & 5 & 3 & 4 & 5 & 3 & 4 & 5\\
    \hline
    32 & 26.36 & 26.18 & 25.46 & 49.53 & 47.64 & 46.96 & 59.06 & 57.10 & 56.47 & 6 & 7 & 7\\
    64 & 27.73 & 27.82 & 27.13 & 50.87 & 50.19 & 49.85 & 61.29 & 59.16 & 59.02 & 5 & 6 & 7\\
    128 & 28.23 & 29.89 & 28.86 & 51.76 & 52.90 & 51.72 & 62.24 & 62.28 & 60.96 & 5 & 5 & 6\\
    256 & \textbf{29.42} & \textbf{30.59} & \textbf{30.23} & \textbf{52.07} & 49.82 & 51.44 & \textbf{60.51} & \textbf{61.73} & \textbf{62.58} & \textbf{4} & \textbf{4} & \textbf{5}\\
    512 & 28.26 & 28.14 & 27.06 & 50.97 & 50.54 & 49.27 & 59.26 & 59.94 & 58.52 & 5 & 5 & 6\\
    \hline
  \end{tabular}
  \end{center}
  \label{tab:goodnews_comparisons}
\end{table*}

\section{Input text length ablation} \label{text_ablation_supp}
To evaluate the importance of the number of input word tokens in retrieving the most relevant image set, we conduct an ablation study over the length of the input text sequences for the best-performing MIL-SIM approach and report the results in Table~\ref{tab:text_len_ablation}. In our experiments, we use both article-level and sentence-level alignment objectives. In the former objective, we begin from the start of an article and limit the number of words to the desired lengths. Since the pretrained CLIP model accepts a maximum of 77 word tokens, we modify the original model by zero-padding the pretrained positional embeddings for the additional word positions and finetune the entire set of positional embeddings during training. 

In the latter objective, we split each article into sentences and limit the number of sentences by the selected number of words. In Table~\ref{tab:goodnews_comparisons},  we observe that using more input words generally helps to improve retrieval accuracy. 

Interestingly, the best retrieval performance is obtained when 256 input word tokens are used. One possible reason is that there are more redundant sentences, which do not really contribute to the overall semantics of the story. The performance drop when 512 word tokens are used also suggests that a solution to this problem has to be able to better filter out non-salient sentences.

\section{Algorithm for selecting multiple images for text narratives} \label{retrieval_algorithm_supp}
 
More often than not, a journalist seeking to obtain suitable images to visually illustrate their article may not be able to find suitable images that have already been grouped into sets. With this in mind, we present a general algorithm below that allows an author to search for individual candidate images before grouping them and using our finetuned models to select the best set.
\newline
 \begin{algorithmic}[1]
 \State Given a text narrative, extract a set of named entities $E = \{e_1, \cdot \cdot \cdot, e_n\}$
 \State $\text{images} \gets \text{set()}$
 \For{\texttt{k in range(n)}}
     \State \texttt{look up a suitable image in an image database using $e_k$}
     \State \text{images.append($I_k$) where $I_k$ is the looked-up image}
 \EndFor
 \State Generate all possible combinations C of images of size X  where X is the desired number of images
 \State $ max\_similarity \gets \inf$
 \State $ \text{best set} \gets None$
 \For{\texttt{k in range(C)}}
     \State compute similarity between text representation and image set representations $score$
     \If{$score > max\_similarity$} 
       \State $max\_similarity = score$
       \State $\text{best set} \gets C_k$
     \EndIf
 \EndFor
 \State return \text{best set}
 \label{img-set-algorithm}
 \end{algorithmic}
 
 \section{Retrieval results using the image search algorithm} 
We present visual examples of the top-ranked image sets for randomly selected articles from the GoodNews dataset using the proposed image set search algorithm and the finetuned MIL-SIM model. For each detected named entity in an article, we search for suitable images using Google Search and select the top 5 results as candidates images. To provide a basis of comparison, we put ourselves in the shoes of a journalist and create image sets for the selected articles using a naive approach. Specifically, we randomly select 5 named entities present in each article and select the top-ranked image returned by the Google Search API for each named entity. While this is a simplistic alternative approach, it can be easily adopted by any journalist without considering one's aesthetic preferences. Compared to these randomly selected image sets, the image sets selected by the MIL-SIM model appear to be more visually descriptive. It is possible that the image sets may be more diverse if a user increases the number of candidate images for each query named entity. However, there is a trade-off between increasing the diversity of the image sets and inference time since the time required to compute the combinations of images increases significantly with more candidate images.

Finally, we conduct a qualitative analysis of the correspondences between individual images in a set and specific sentences in an article. Recall that the MIL-SIM approach relies on the assumption that each image should be related to at least one sentence in an article, even if their relationship is loose and illustrative at best rather than literal. We compute the similarity scores between each sentence in an article and the image set selected by the MIL-SIM model. Each image and its best matching sentence are outlined and highlighted in the same color.

\def\ourfigwidth{0.8}

\begin{figure}[bht]
\begin{center}
\includegraphics[width=\ourfigwidth\linewidth]{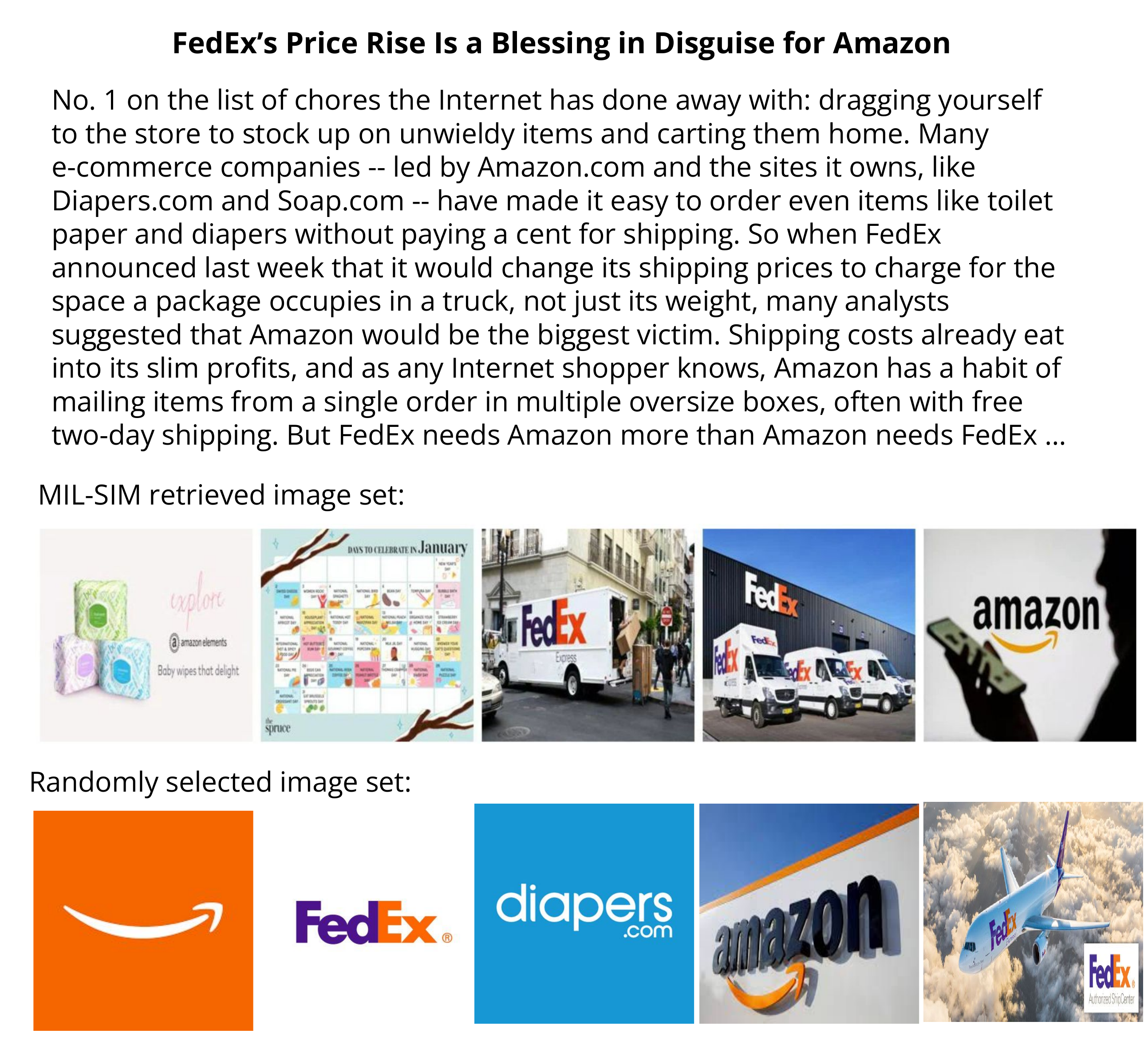}
\end{center}
 \caption{Example of retrieved image sets given a query article using our algorithm.}
\end{figure}

\begin{figure}[htb]
\begin{center}
\includegraphics[width=\ourfigwidth\linewidth]{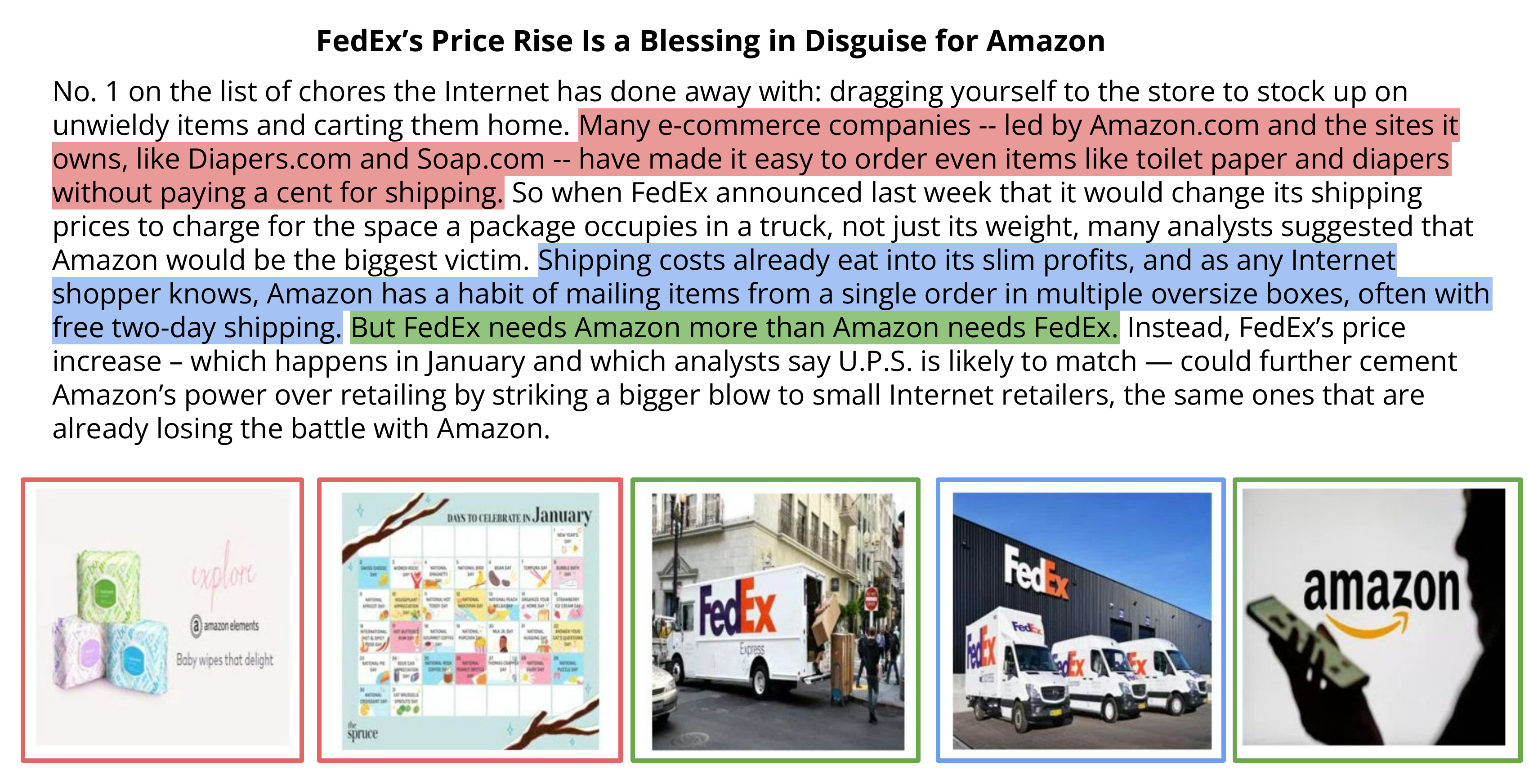}
\end{center}
 \caption{Example of the most relevant sentences for each image in the top-ranked image set, as determined by the MIL-SIM model.}
\end{figure}


\begin{figure}[bth]
\begin{center}
\includegraphics[width=\ourfigwidth\linewidth]{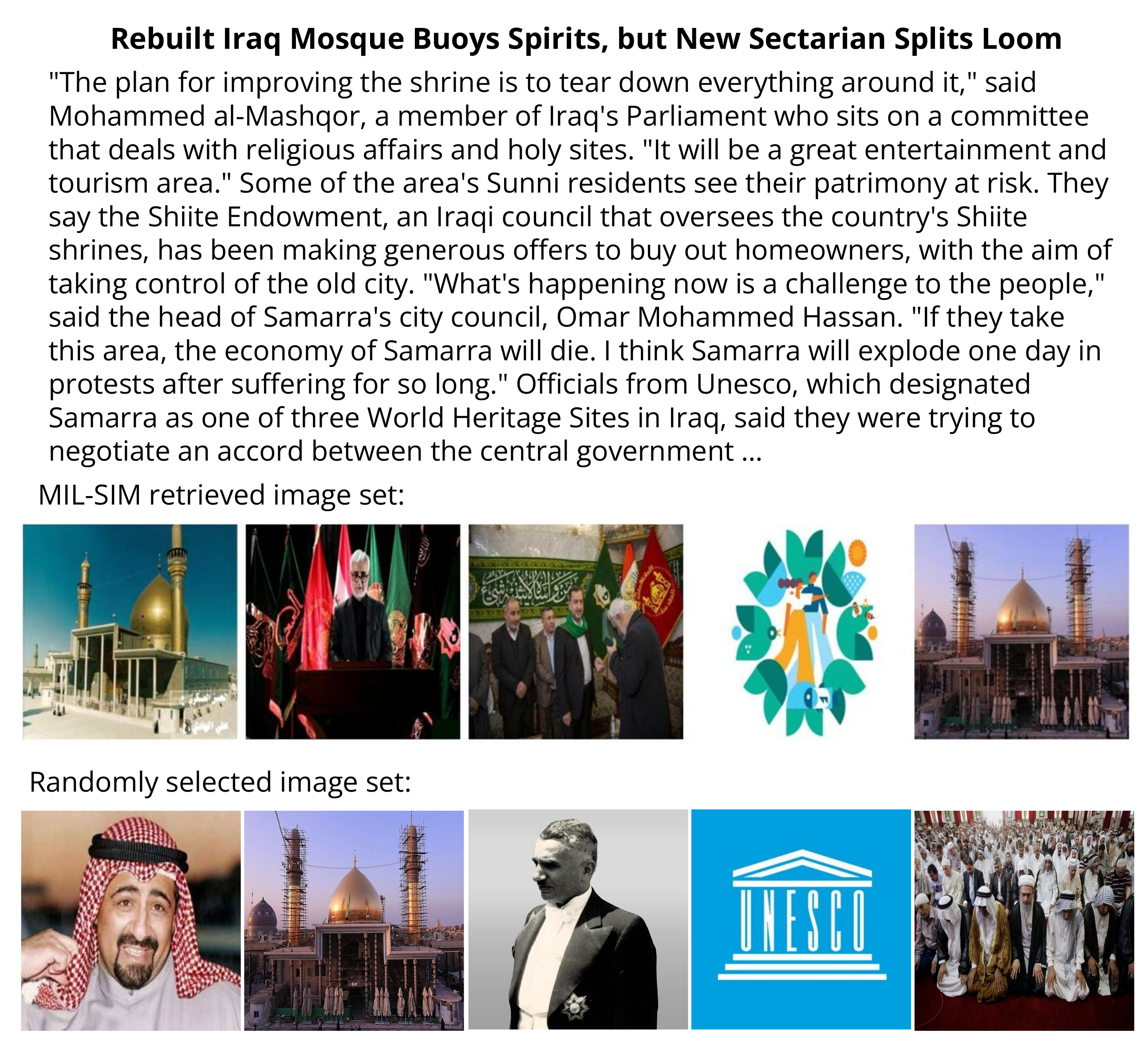}
\end{center}
 \caption{Example of retrieved image sets given a query article using our algorithm.}
\end{figure}

\begin{figure}[thb]
\begin{center}
\includegraphics[width=\ourfigwidth\linewidth]{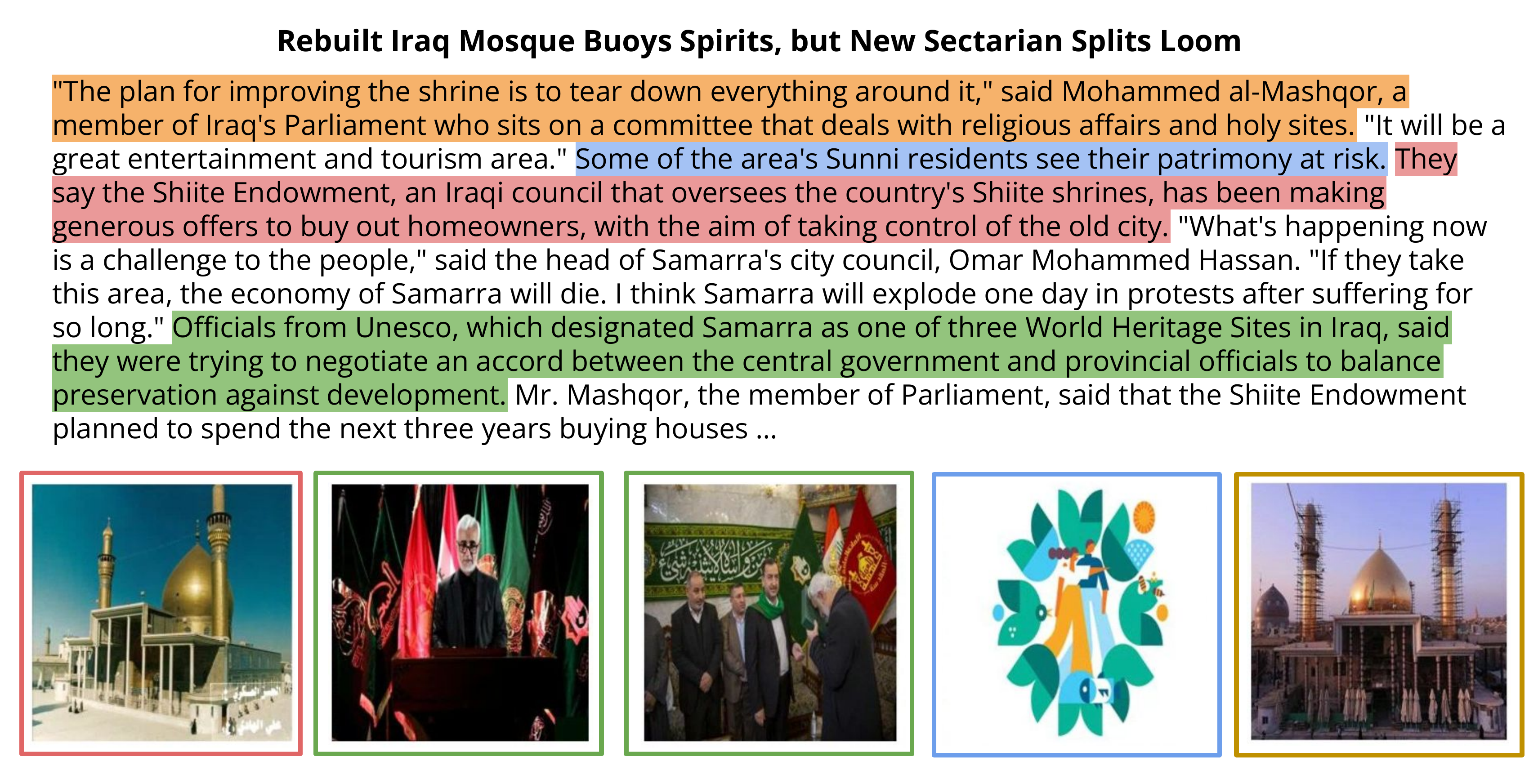}
\end{center}
 \caption{Example of the most relevant sentences for each image in the top-ranked image set, as determined by the MIL-SIM model.}
\end{figure}

\begin{figure}[htb]
\begin{center}
\includegraphics[width=\ourfigwidth\linewidth]{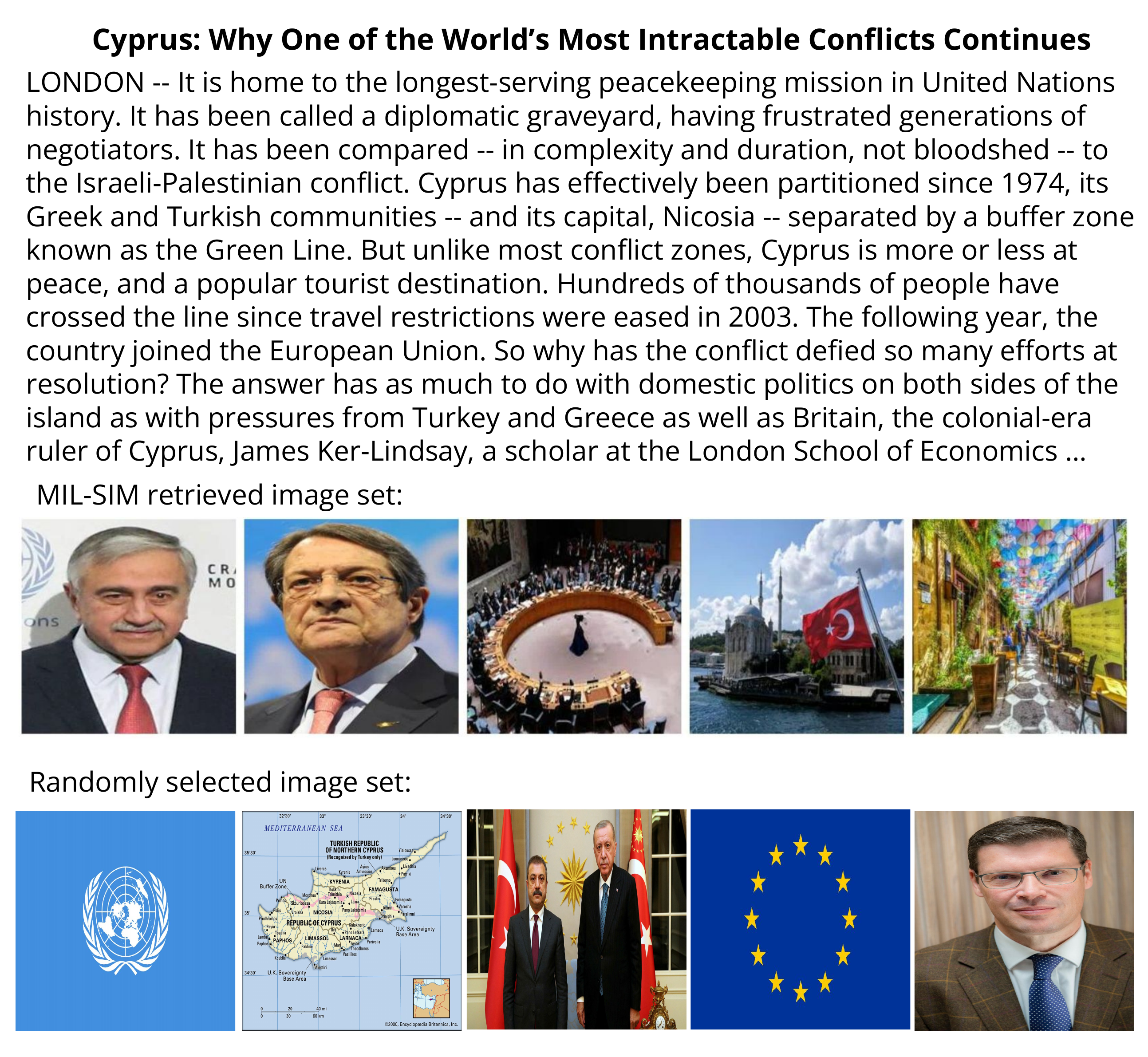}
\end{center}
 \caption{Example of retrieved image sets given a query article using our algorithm.}
\end{figure}

\begin{figure}[t]
\begin{center}
\includegraphics[width=\ourfigwidth\linewidth]{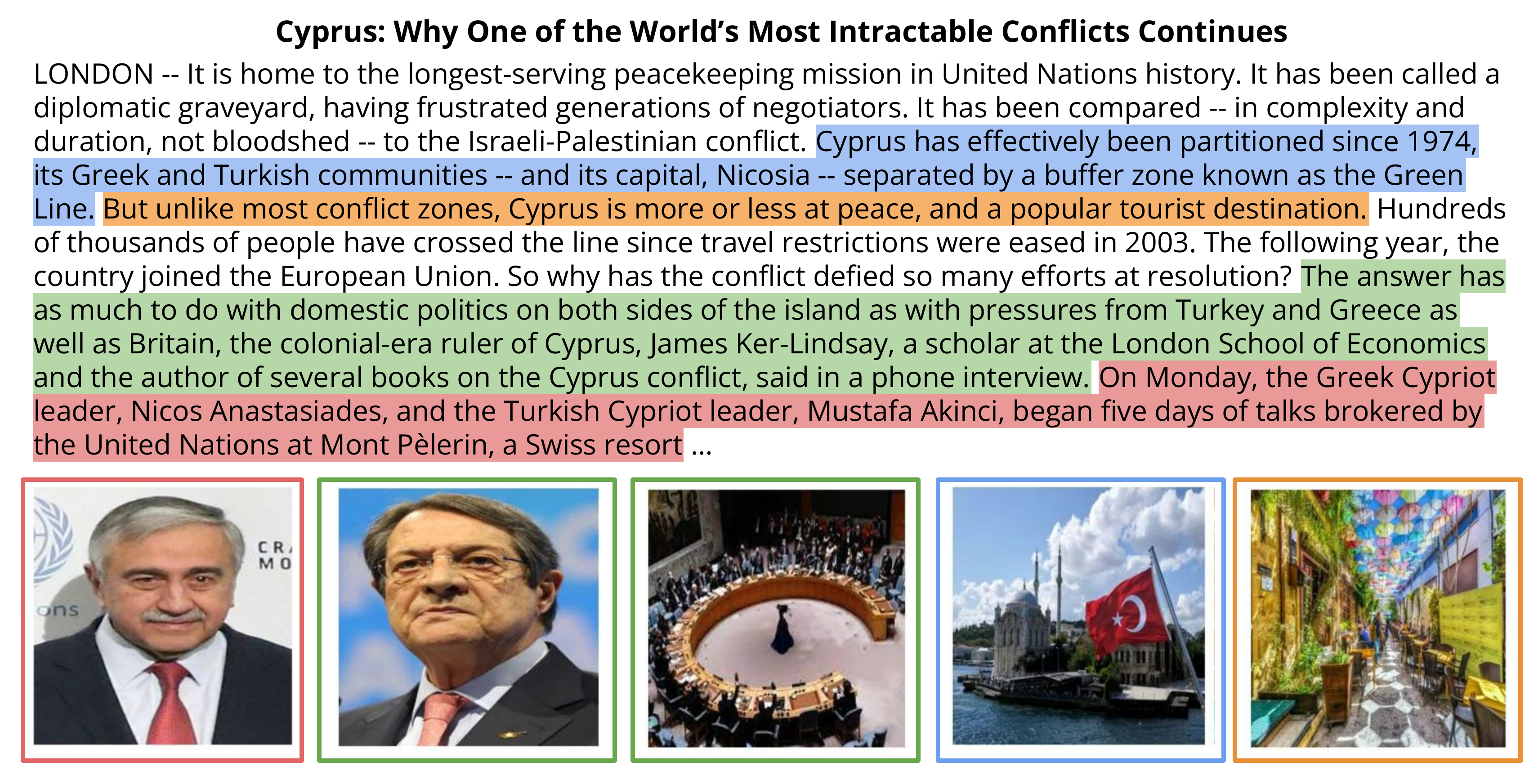}
\end{center}
 \caption{Example of the most relevant sentences for each image in the top-ranked image set, as determined by the MIL-SIM model.}
\end{figure}

\begin{figure}[h]
\begin{center}
\includegraphics[width=\ourfigwidth\linewidth]{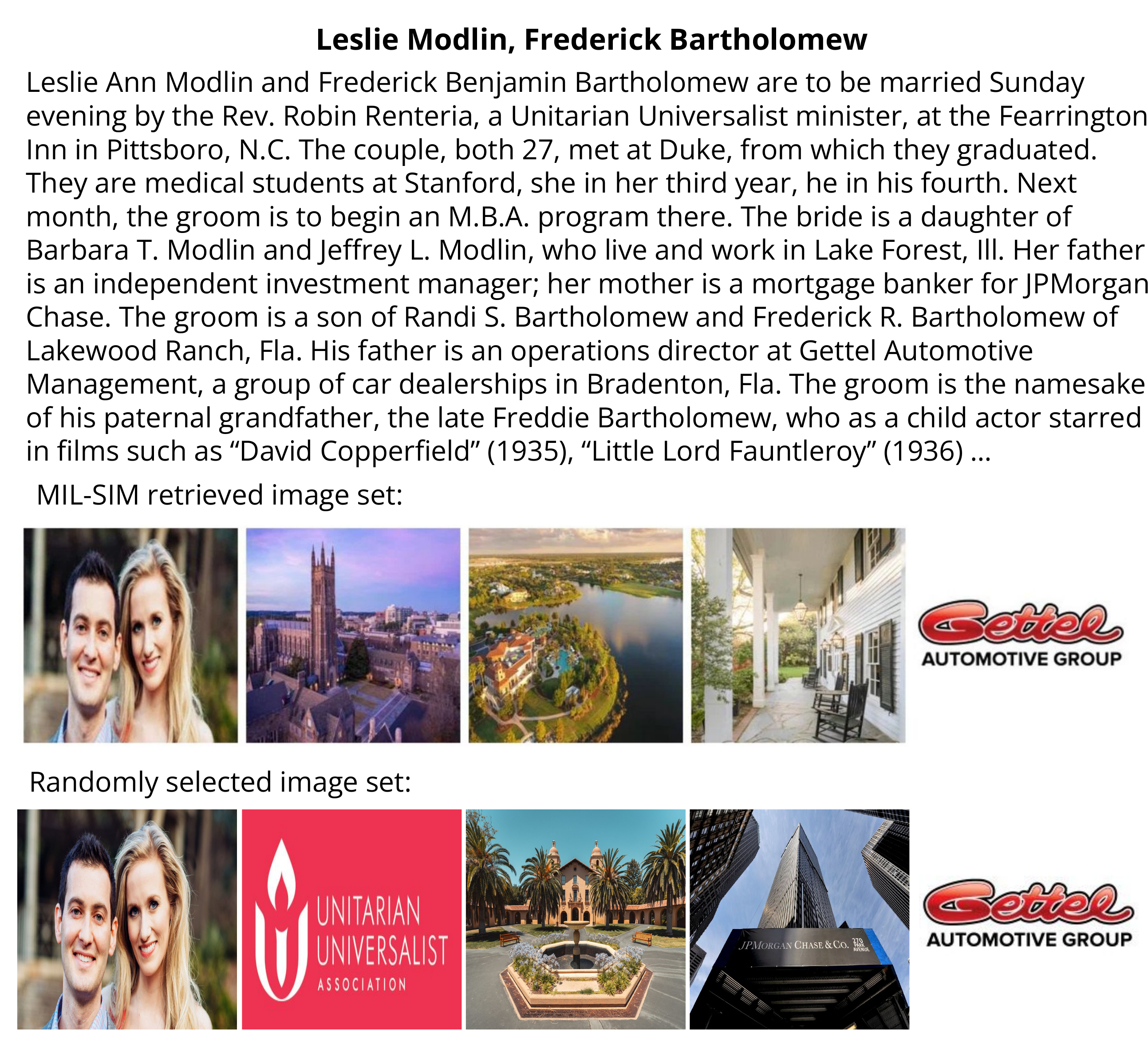}
\end{center}
 \caption{Example of retrieved image sets given a query article using our algorithm.}
\end{figure}

\begin{figure}[t]
\begin{center}
\includegraphics[width=\ourfigwidth\linewidth]{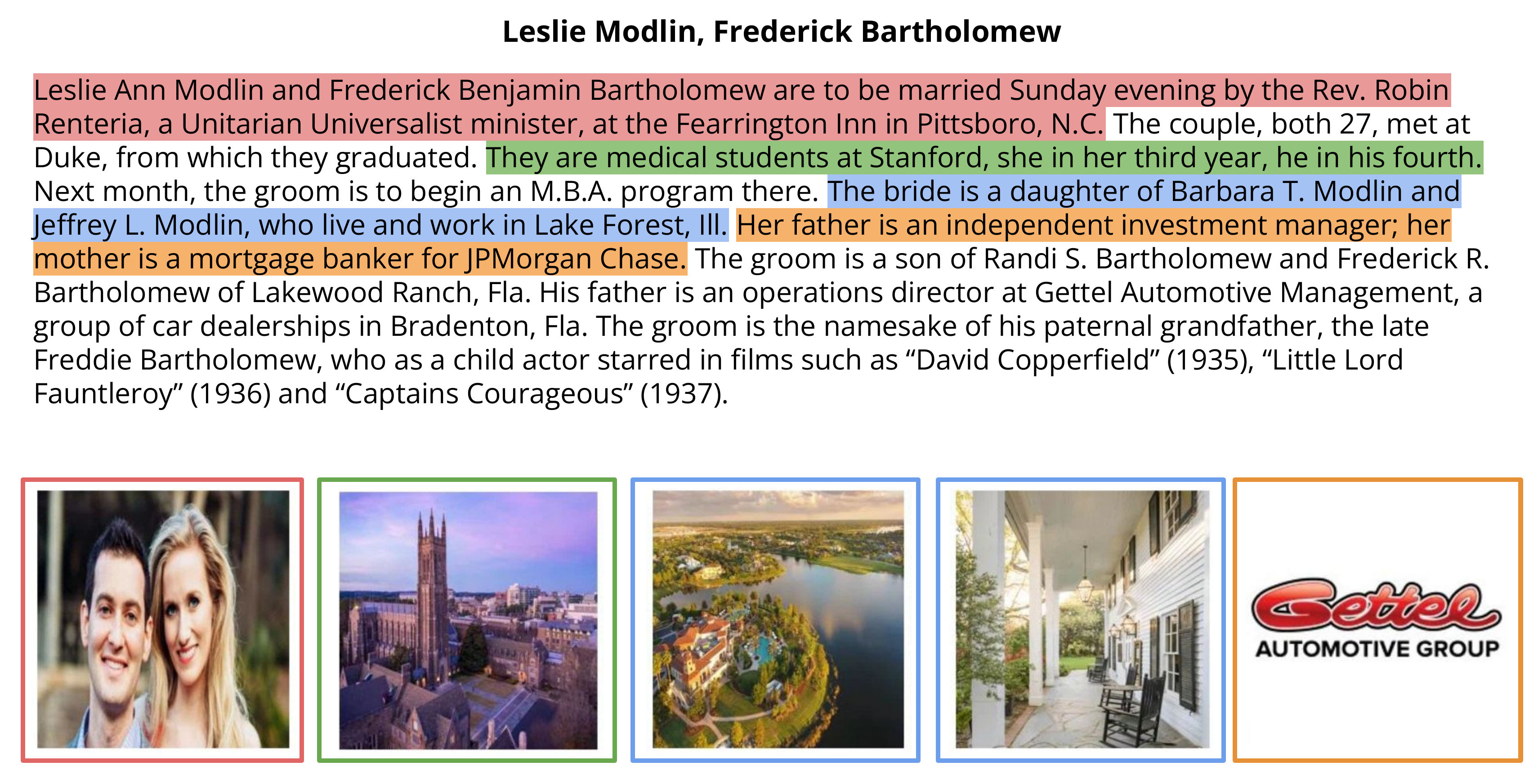}
\end{center}
 \caption{Example of the most relevant sentences for each image in the top-ranked image set, as determined by the MIL-SIM model.}
\end{figure}

\begin{figure}[h]
\begin{center}
\includegraphics[width=\ourfigwidth\linewidth]{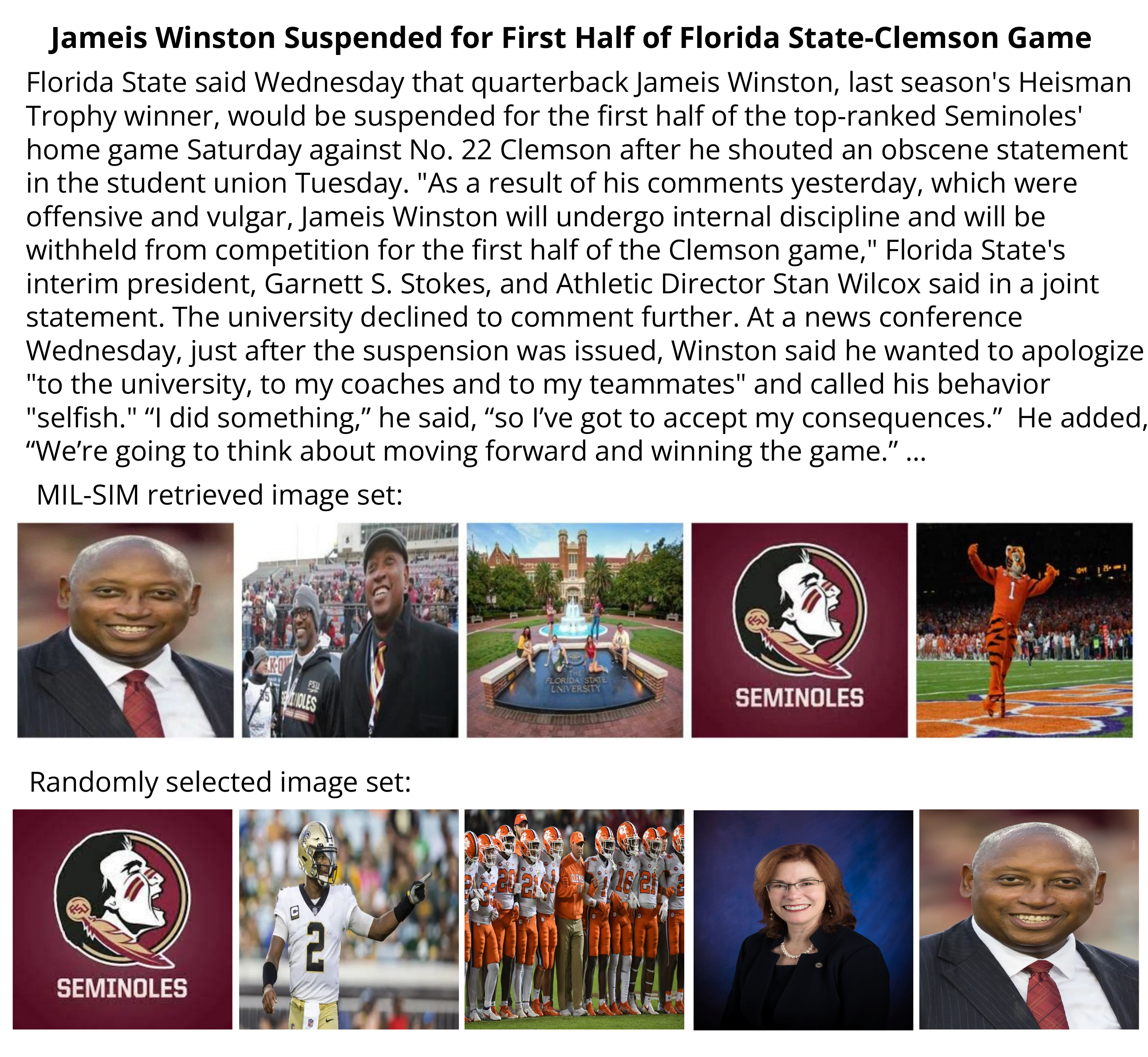}
\end{center}
 \caption{Example of retrieved image sets given a query article using our algorithm.}
\end{figure}

\begin{figure}[t]
\begin{center}
\includegraphics[width=\ourfigwidth\linewidth]{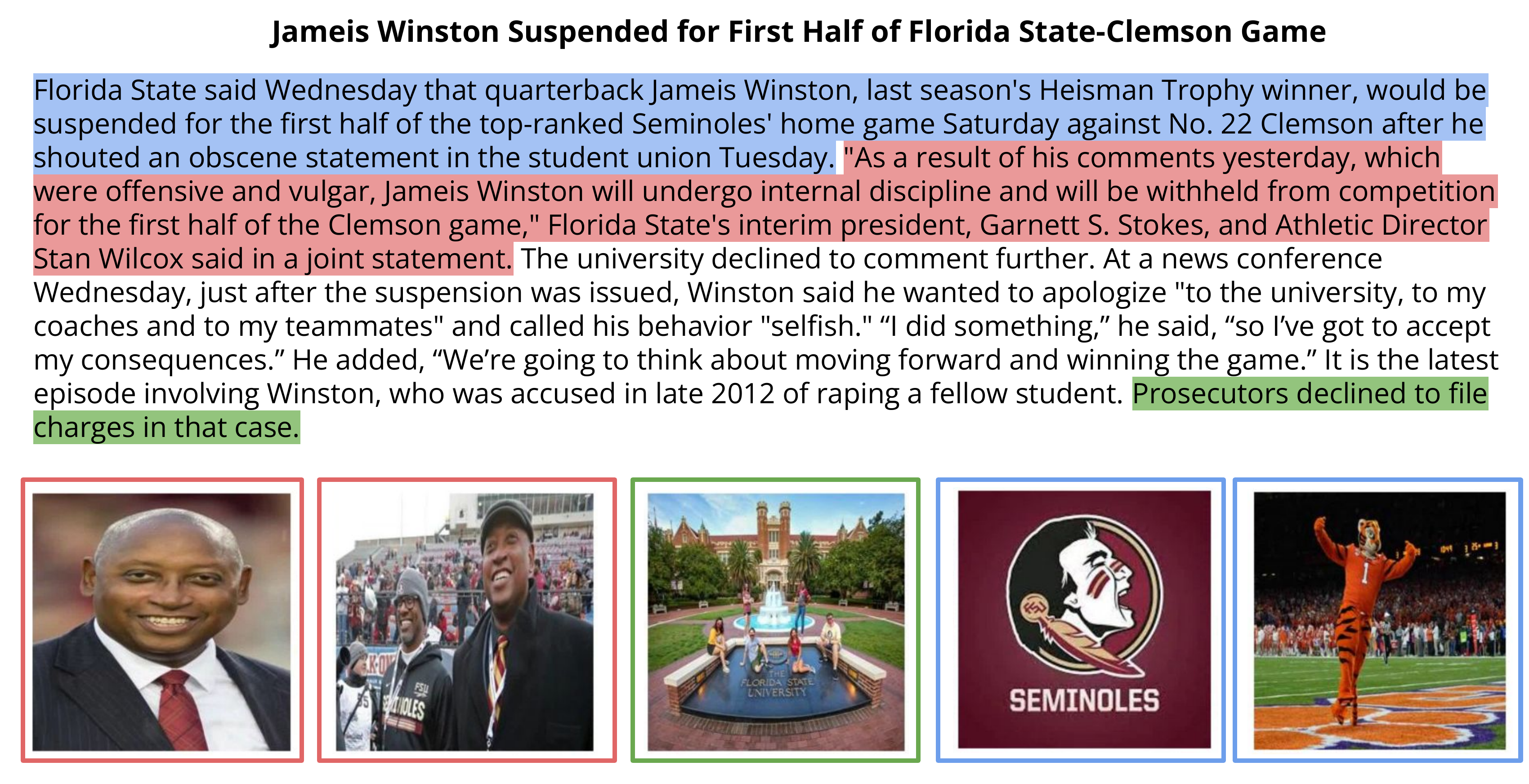}
\end{center}
 \caption{Example of the most relevant sentences for each image in the top-ranked image set, as determined by the MIL-SIM model.}
\end{figure}  

\clearpage

\subsubsection{Image set retrievals for articles} \label{retrieval_visualizations_supp}
 
 \subsection{Additional qualitative retrieval results}
In this section, we provide additional qualitative retrieval results obtained by our best-performing MIL-SIM approach on our test splits of the \dataset and GoodNews datasets. For each query article, we show the ground-truth set of corresponding images as well as the top five retrieved image sets. In the visualizations, we also provide the corresponding article titles that correspond to the retrieved image sets. The ground-truth and incorrect image sets are also outlined in green and red boxes, respectively. The retrieved image sets for a query article are ordered from top to bottom.

\begin{figure*}[t]
\begin{center}
\includegraphics[width=\linewidth, height=18cm]{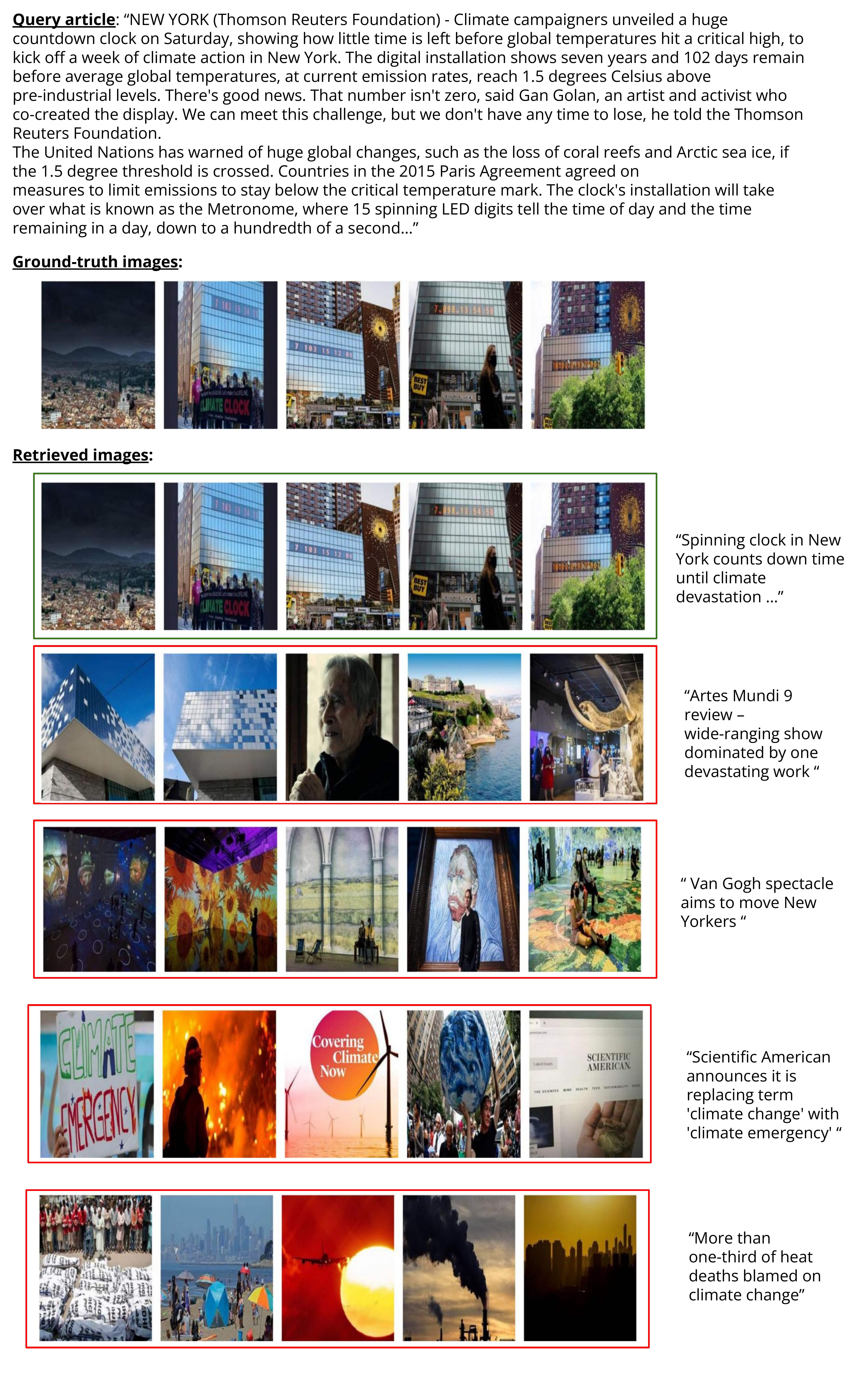}
\vspace{-9cm}
\end{center}
   \caption{Correct retrievals on the test split of the \dataset dataset.}
\label{fig:supp_goodnews_qualitative}
\vspace{\fullpagefigurepad}
\end{figure*}

\begin{figure*}[t]
\begin{center}
\includegraphics[width=\linewidth, height=18cm]{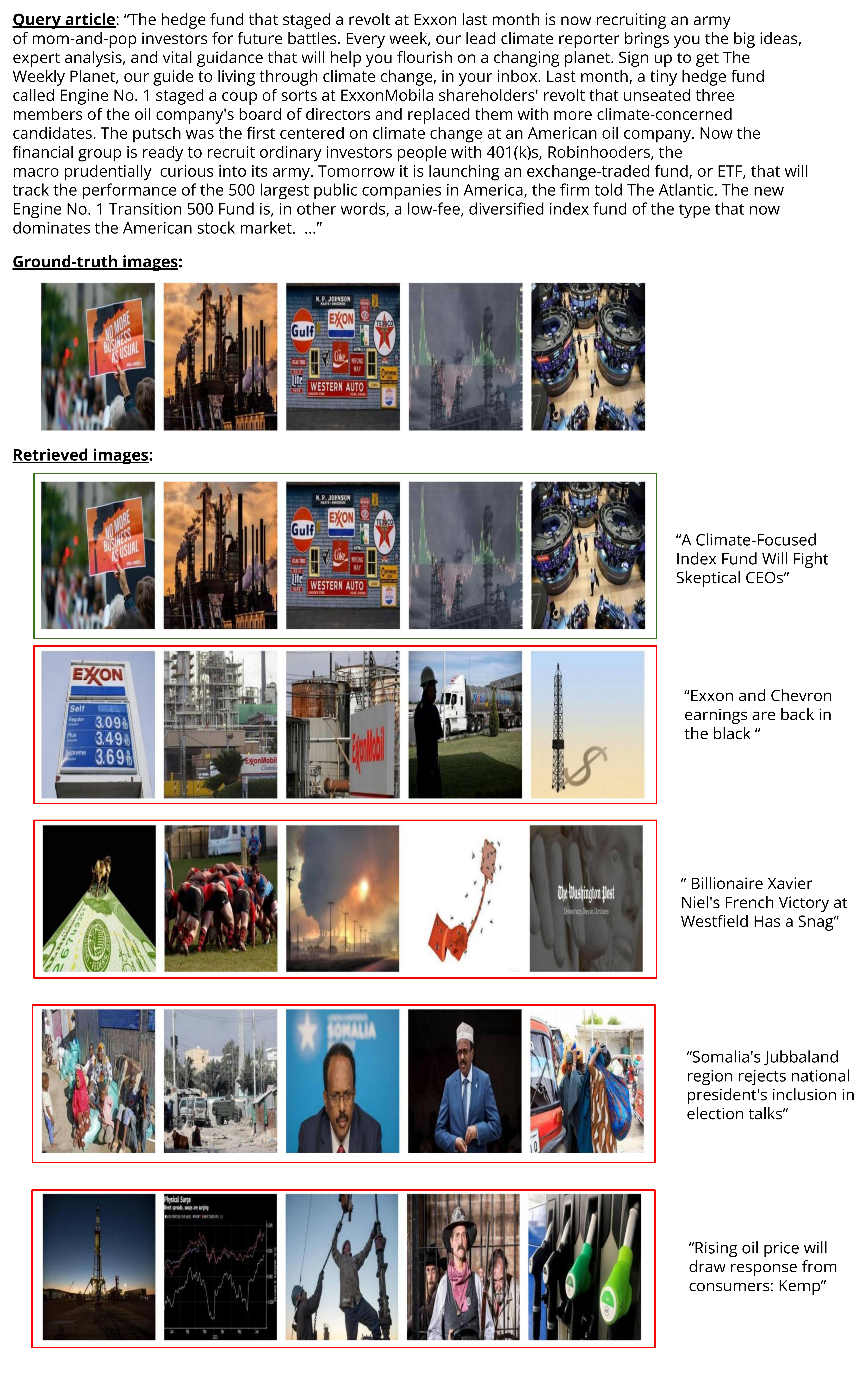}
\vspace{-9cm}
\end{center}
   \caption{Correct retrievals on the test split of the \dataset dataset.}
\label{fig:supp_goodnews_qualitative}
\vspace{\fullpagefigurepad}
\end{figure*}

\begin{figure*}[t]
\begin{center}
\includegraphics[width=\linewidth, height=18cm]{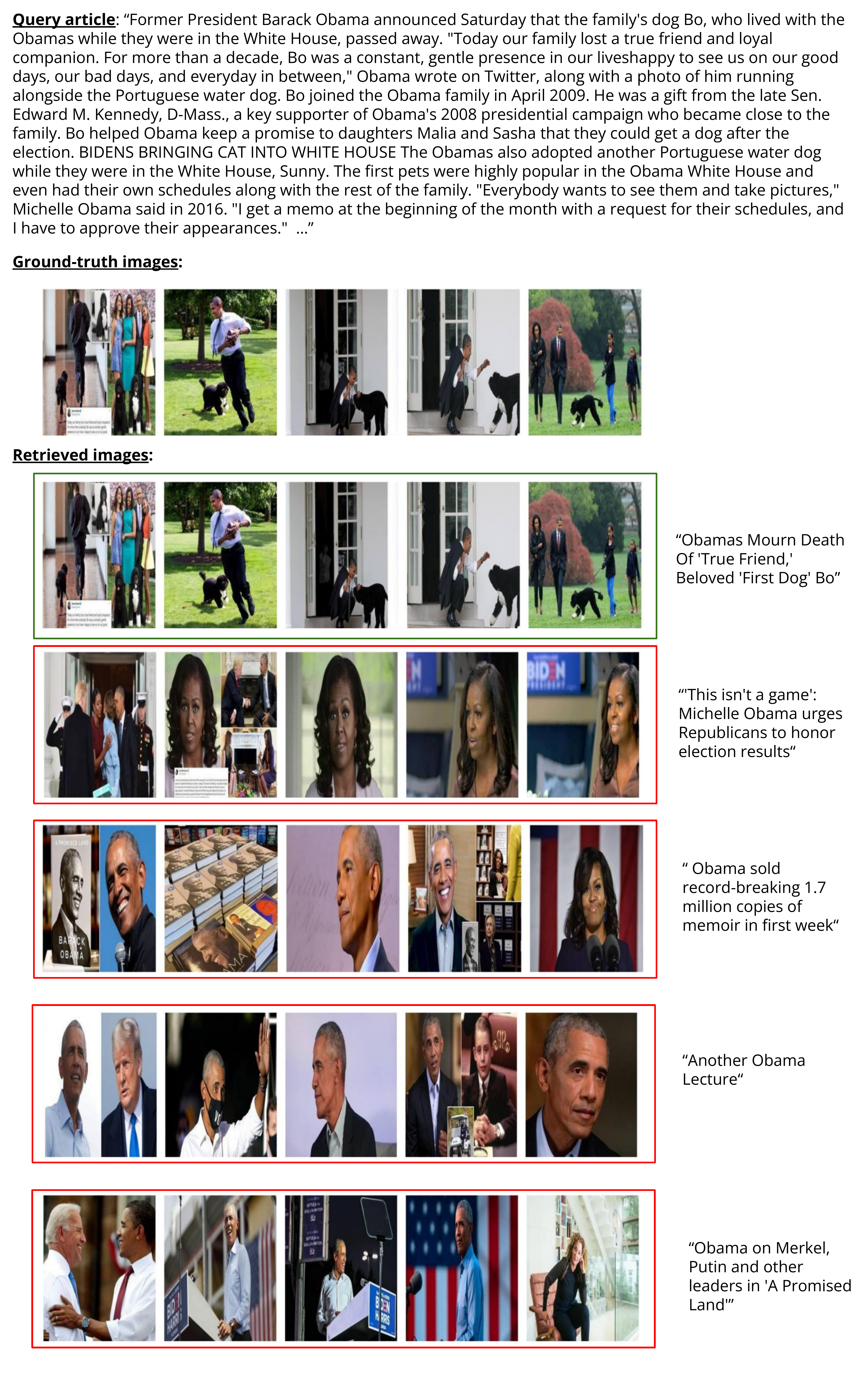}
\vspace{-9cm}
\end{center}
   \caption{Correct retrievals on the test split of the \dataset dataset.}
\label{fig:supp_goodnews_qualitative}
\vspace{\fullpagefigurepad}
\end{figure*}

\begin{figure*}[t]
\begin{center}
\includegraphics[width=\linewidth, height=18cm]{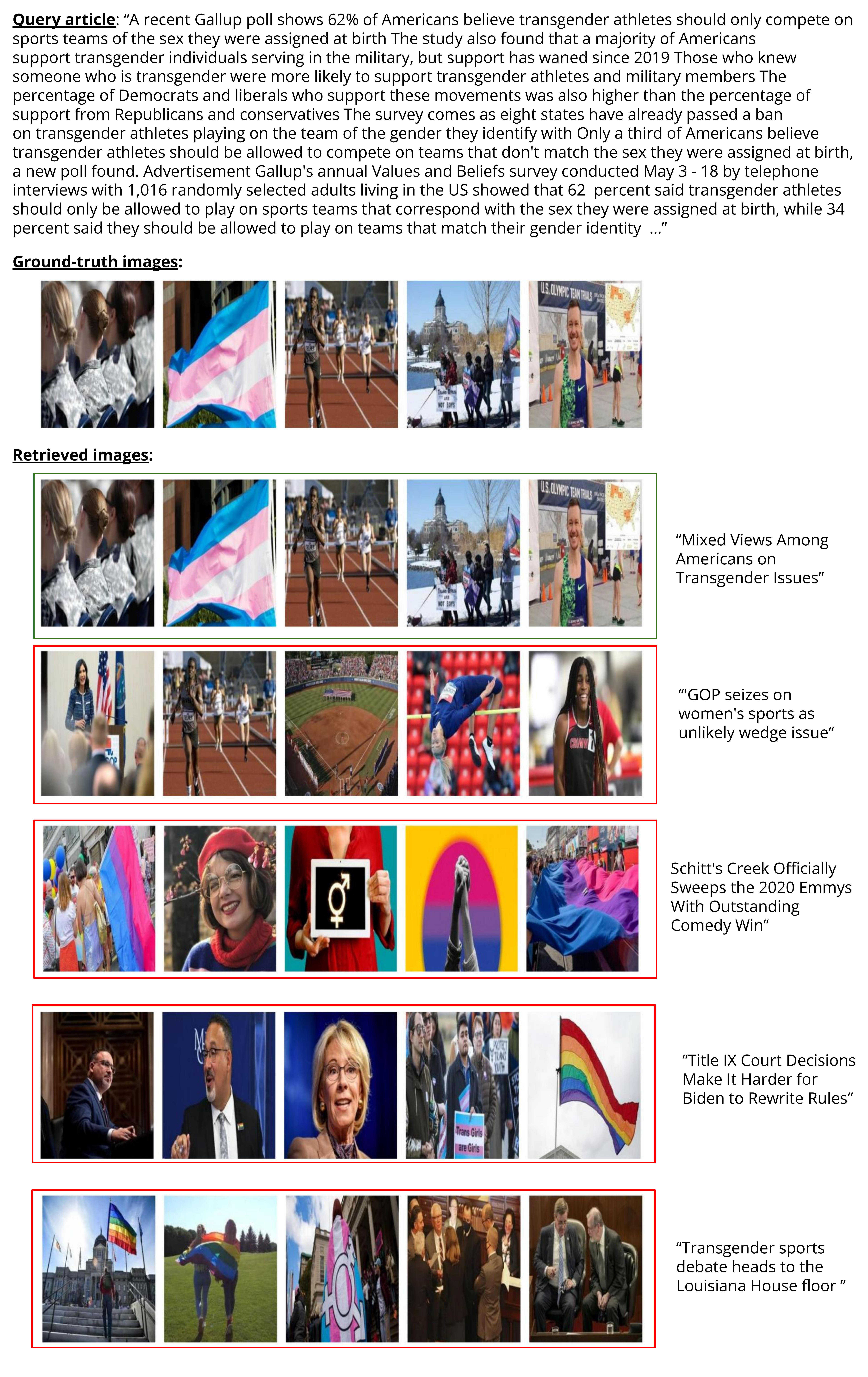}
\vspace{-9cm}
\end{center}
   \caption{Correct retrievals on the test split of the \dataset dataset.}
\label{fig:supp_goodnews_qualitative}
\vspace{\fullpagefigurepad}
\end{figure*}

\begin{figure*}[t]
\begin{center}
\includegraphics[width=\linewidth, height=18cm]{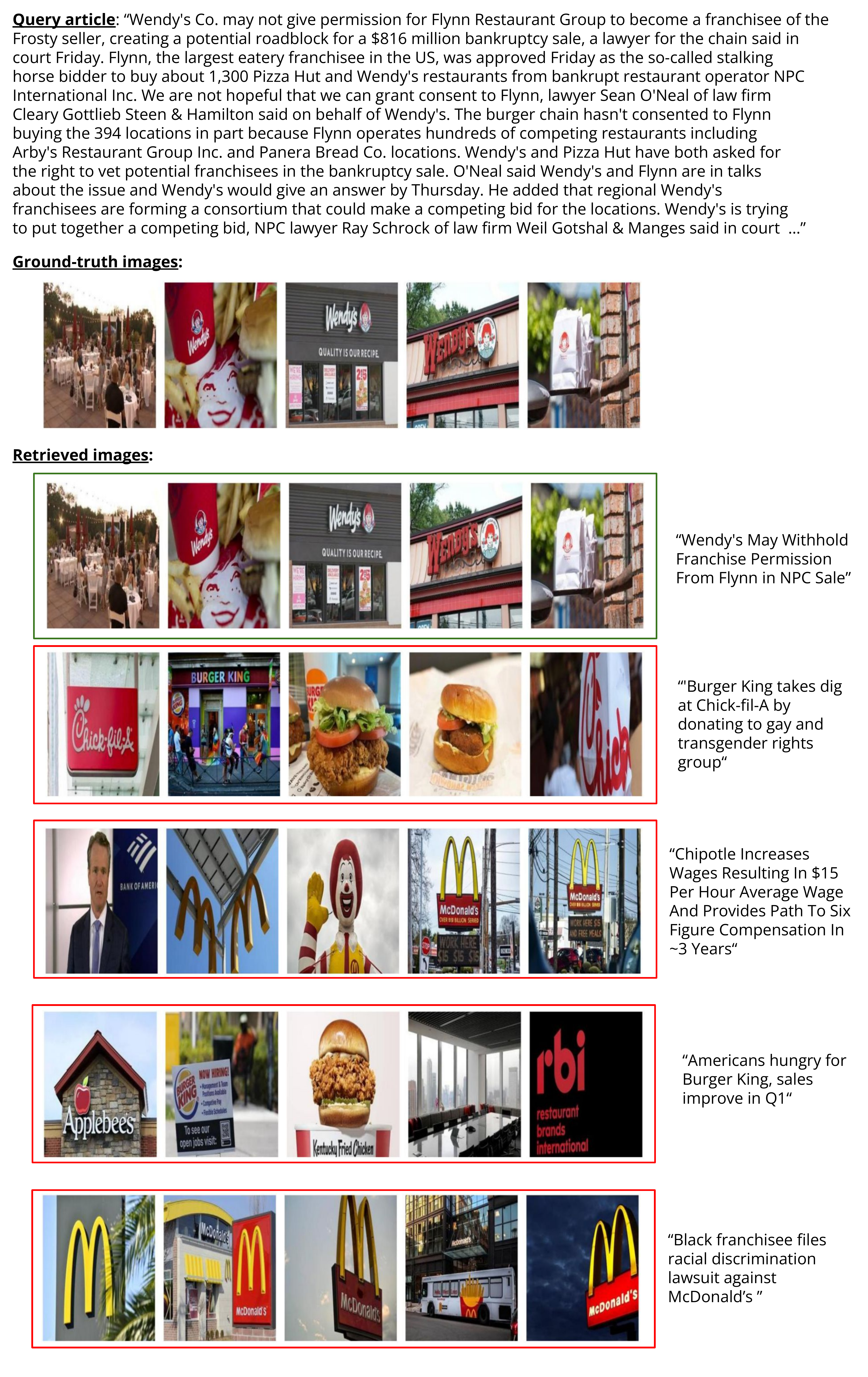}
\vspace{-9cm}
\end{center}
   \caption{Correct retrievals on the test split of the \dataset dataset.}
\label{fig:supp_goodnews_qualitative}
\vspace{\fullpagefigurepad}
\end{figure*}

\begin{figure*}[t]
\begin{center}
\includegraphics[width=\linewidth, height=18cm]{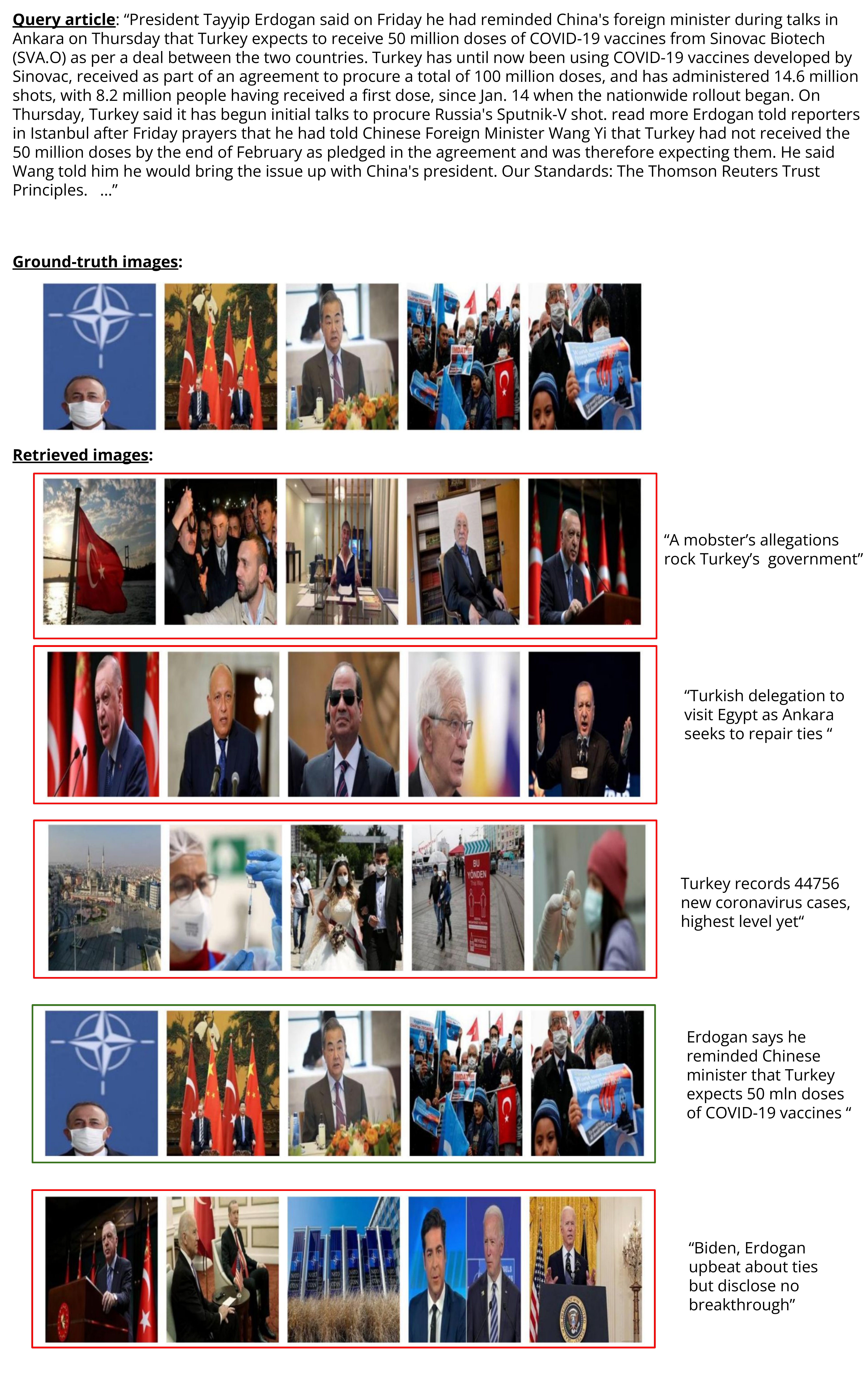}
\vspace{-9cm}
\end{center}
   \caption{Incorrect retrievals on the test split of the \dataset dataset.}
\label{fig:supp_goodnews_qualitative}
\vspace{\fullpagefigurepad}
\end{figure*}

\begin{figure*}[t]
\begin{center}
\includegraphics[width=\linewidth, height=18cm]{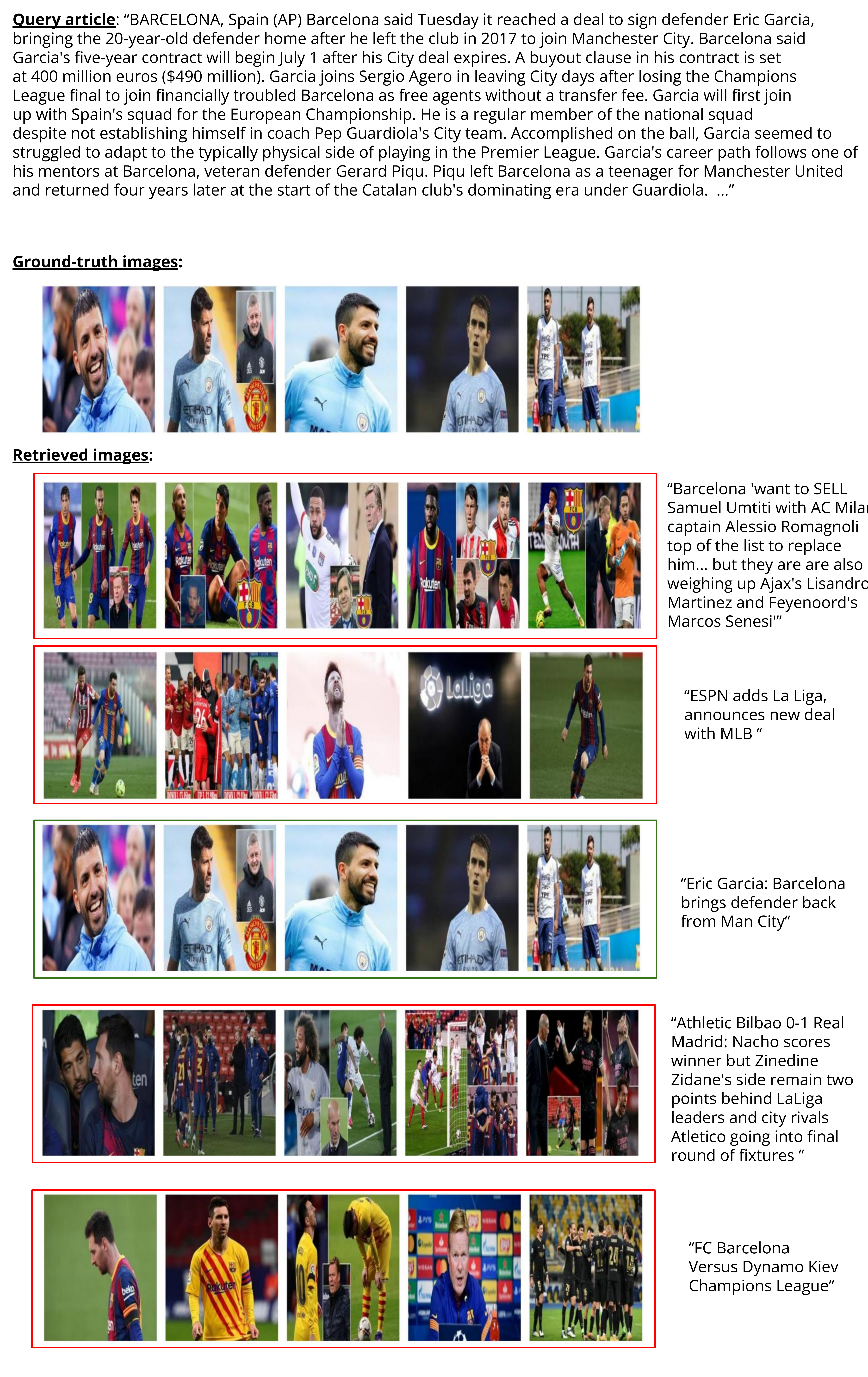}
\vspace{-9cm}
\end{center}
   \caption{Incorrect retrievals on the test split of the \dataset dataset.}
\label{fig:supp_goodnews_qualitative}
\vspace{\fullpagefigurepad}
\end{figure*}

\begin{figure*}[t]
\begin{center}
\includegraphics[width=\linewidth, height=18cm]{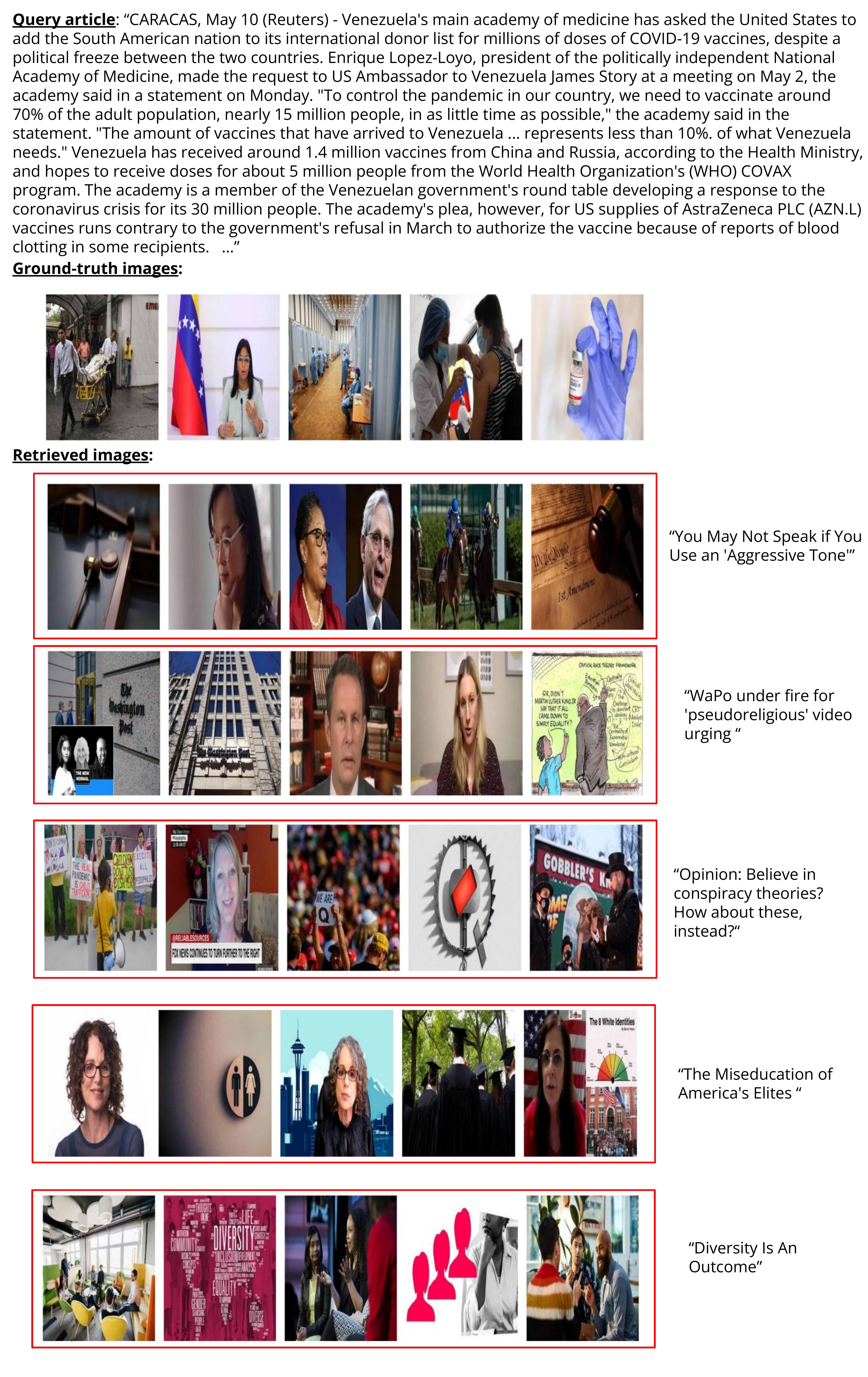}
\vspace{-9cm}
\end{center}
   \caption{Incorrect retrievals on the test split of the \dataset dataset.}
\label{fig:supp_goodnews_qualitative}
\vspace{\fullpagefigurepad}
\end{figure*}

\begin{figure*}[t]
\begin{center}
\includegraphics[width=\linewidth, height=18cm]{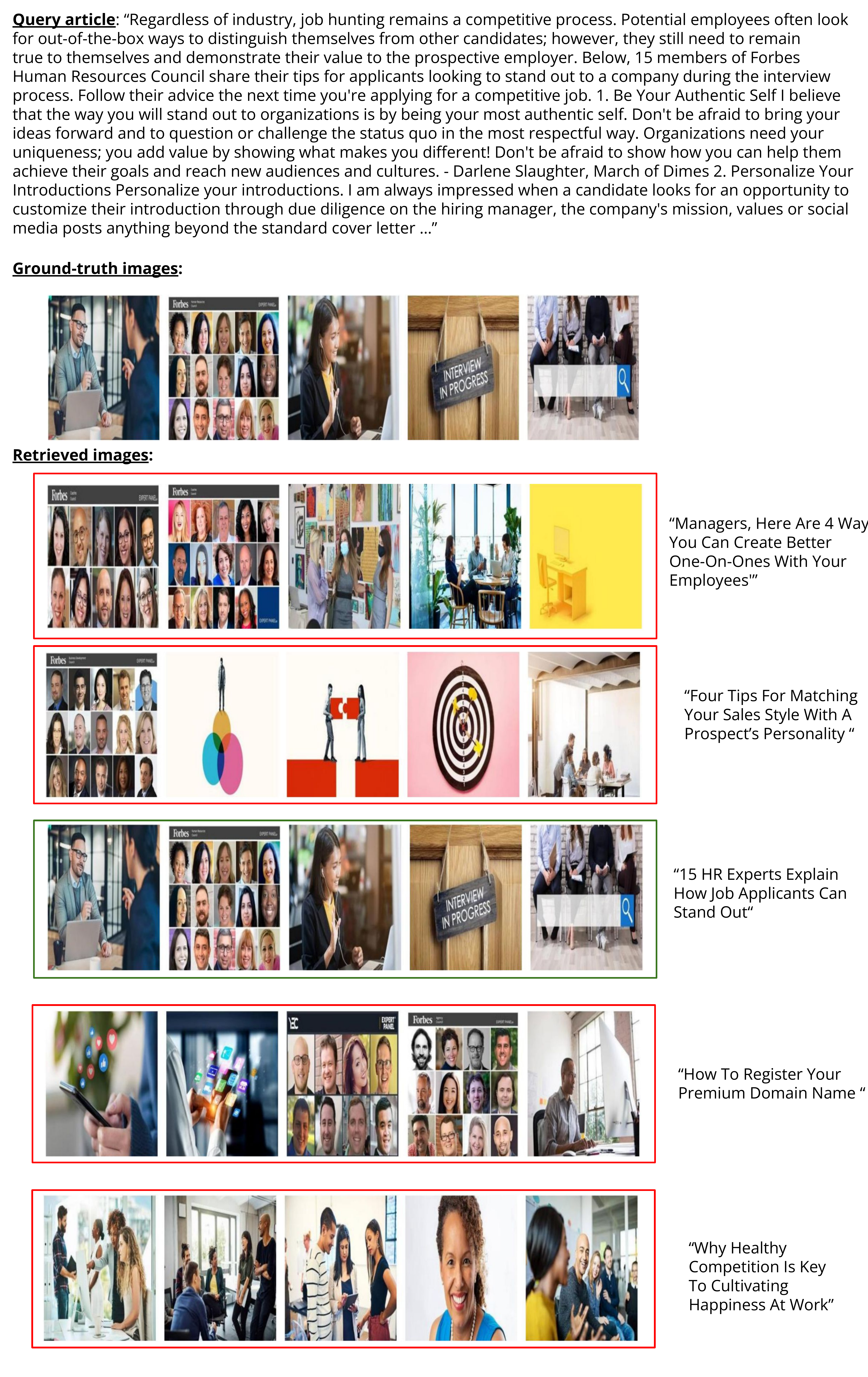}
\vspace{-9cm}
\end{center}
   \caption{Incorrect retrievals on the test split of the \dataset dataset.}
\label{fig:supp_goodnews_qualitative}
\vspace{\fullpagefigurepad}
\end{figure*}

\begin{figure*}[t]
\begin{center}
\includegraphics[width=\linewidth, height=18cm]{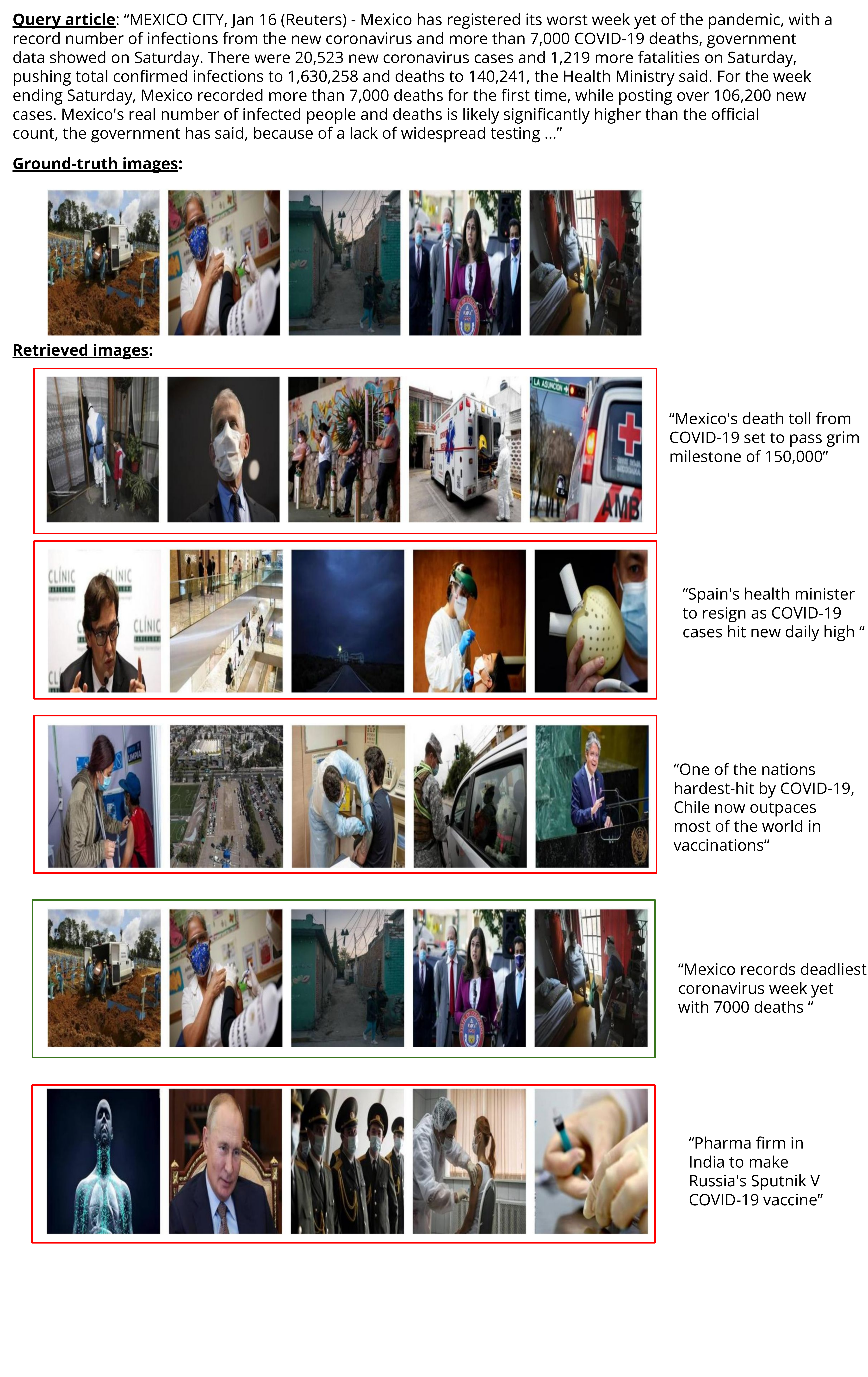}
\vspace{-9cm}
\end{center}
   \caption{Incorrect retrievals on the test split of the \dataset dataset.}
\label{fig:supp_goodnews_qualitative}
\vspace{\fullpagefigurepad}
\end{figure*}


\begin{figure*}[t]
\begin{center}
\includegraphics[width=\linewidth, height=18cm]{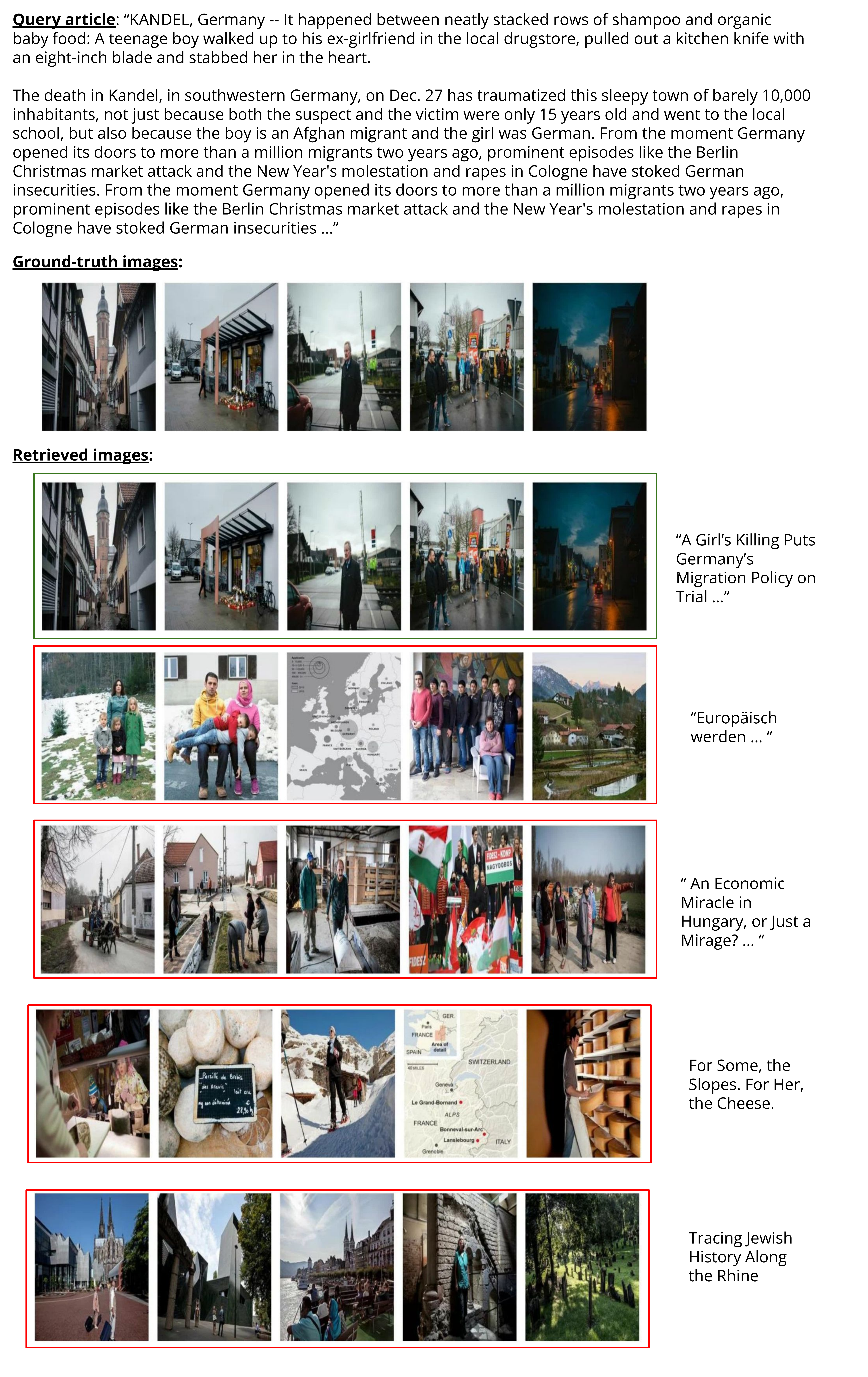}
\vspace{-9cm}
\end{center}
   \caption{Correct retrievals on the GoodNews dataset.}
\label{fig:supp_goodnews_qualitative}
\vspace{\fullpagefigurepad}
\end{figure*}

\begin{figure*}[t]
\begin{center}
\includegraphics[width=\linewidth, height=18cm]{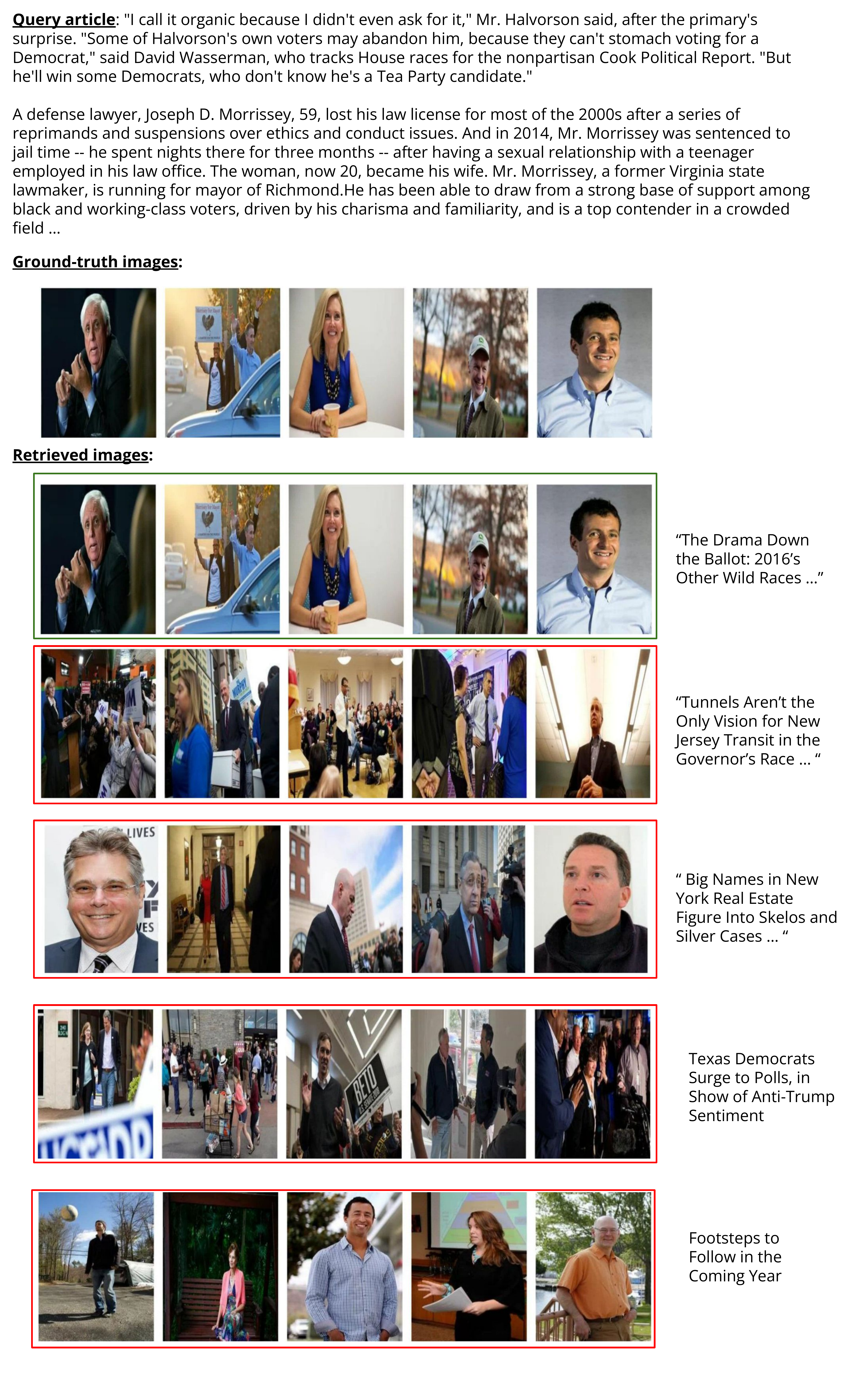}
\vspace{-9cm}
\end{center}
   \caption{Correct retrievals on the GoodNews dataset.}
\label{fig:supp_goodnews_qualitative}
\vspace{\fullpagefigurepad}
\end{figure*}

\begin{figure*}[t]
\begin{center}
\includegraphics[width=\linewidth, height=18cm]{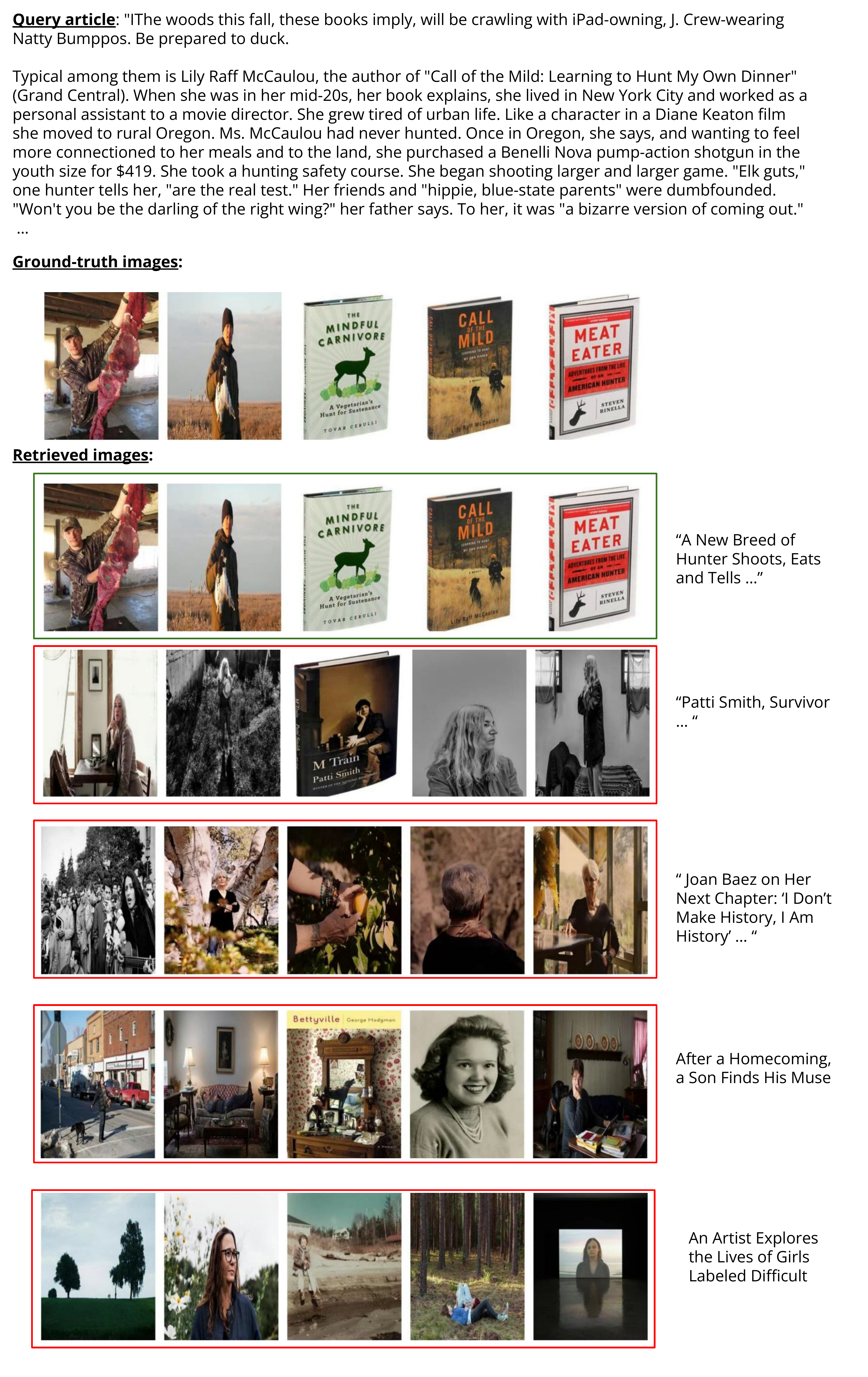}
\vspace{-9cm}
\end{center}
   \caption{Correct retrievals on the GoodNews dataset.}
\label{fig:supp_goodnews_qualitative}
\vspace{\fullpagefigurepad}
\end{figure*}

\begin{figure*}[t]
\begin{center}
\includegraphics[width=\linewidth, height=18cm]{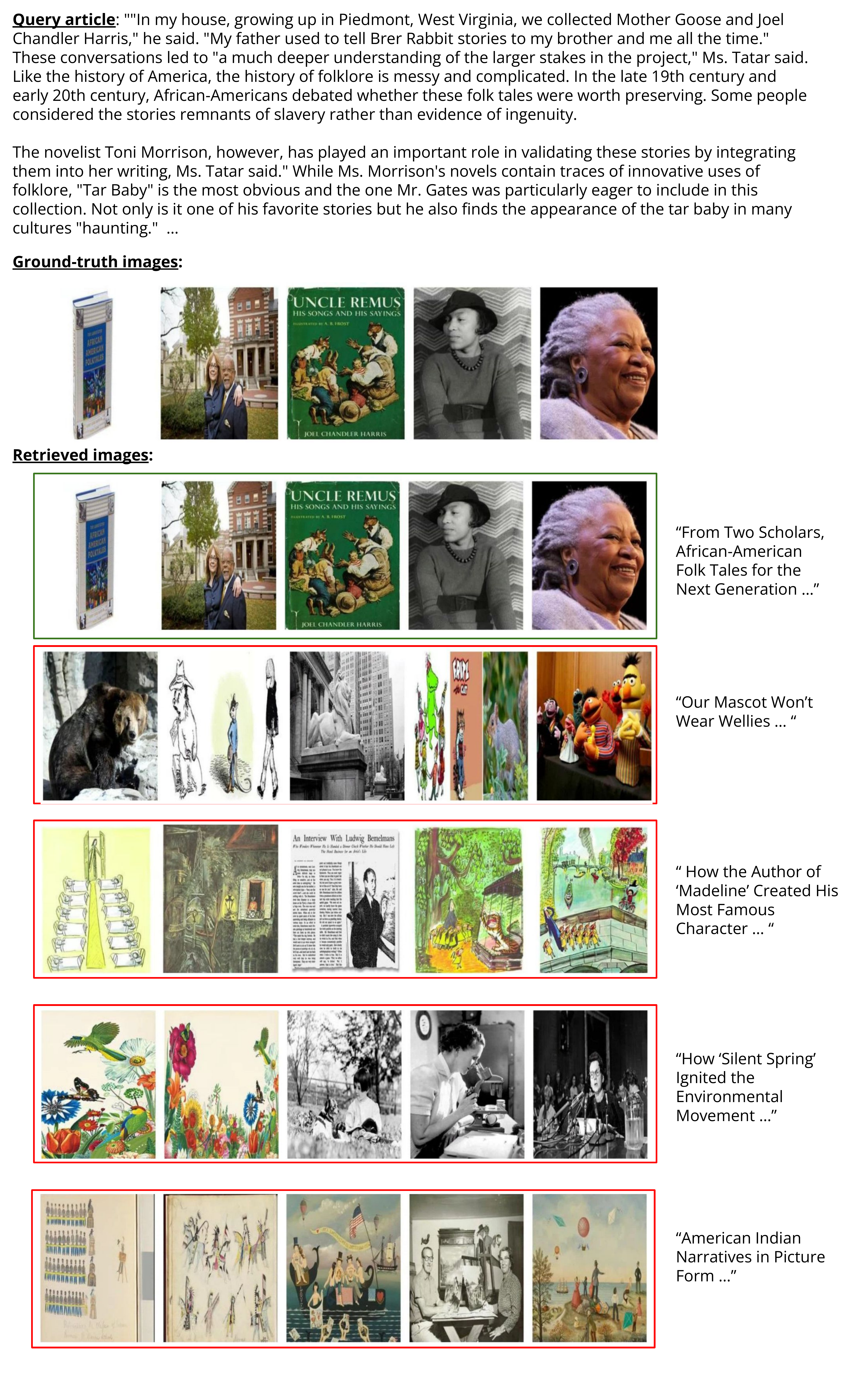}
\vspace{-9cm}
\end{center}
   \caption{Correct retrievals on the GoodNews dataset.}
\label{fig:supp_goodnews_qualitative}
\vspace{\fullpagefigurepad}
\end{figure*}

\begin{figure*}[t]
\begin{center}
\includegraphics[width=\linewidth, height=18cm]{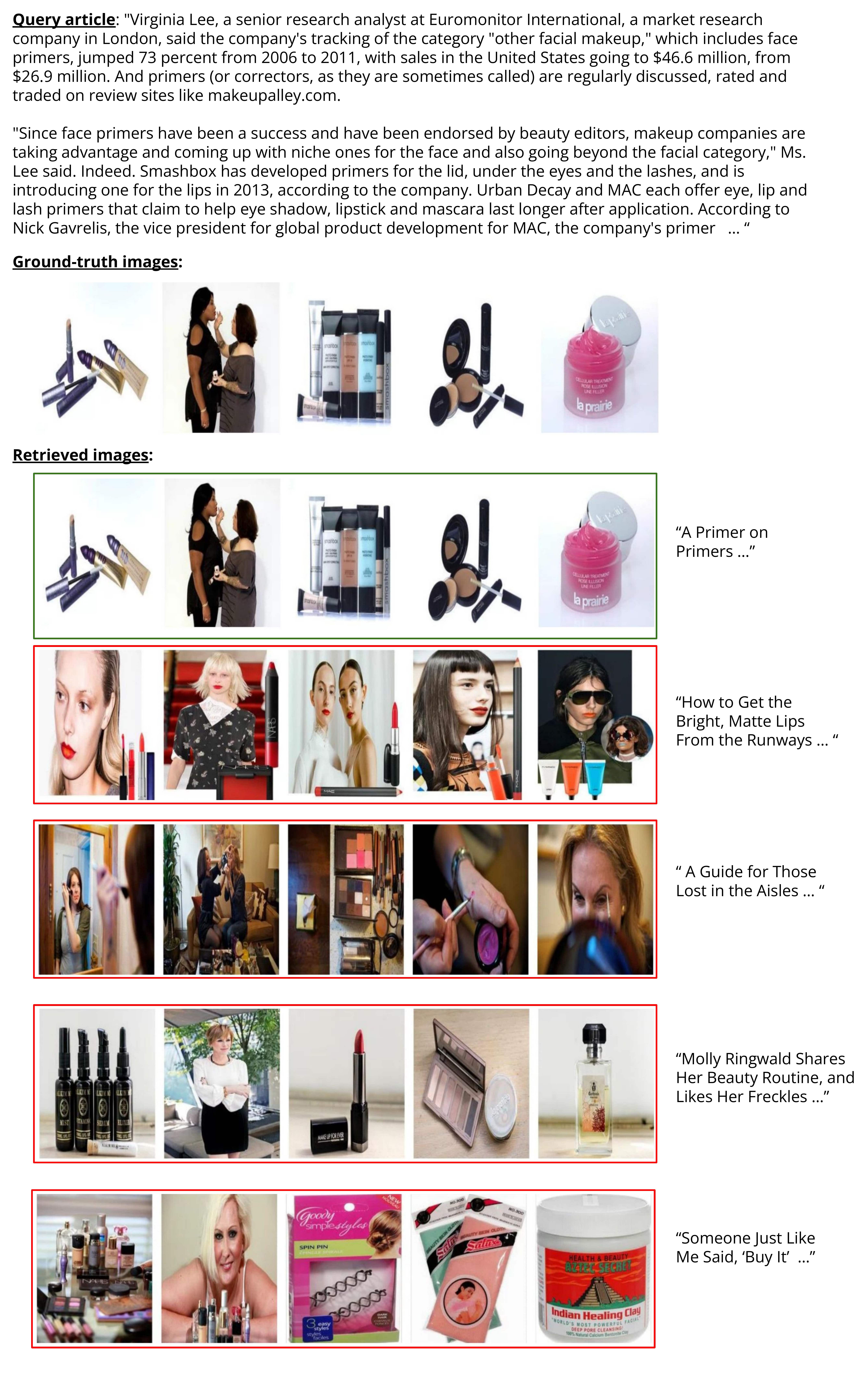}
\vspace{-9cm}
\end{center}
   \caption{Correct retrievals on the GoodNews dataset.}
\label{fig:supp_goodnews_qualitative}
\vspace{\fullpagefigurepad}
\end{figure*}

\begin{figure*}[t]
\begin{center}
\includegraphics[width=\linewidth, height=18cm]{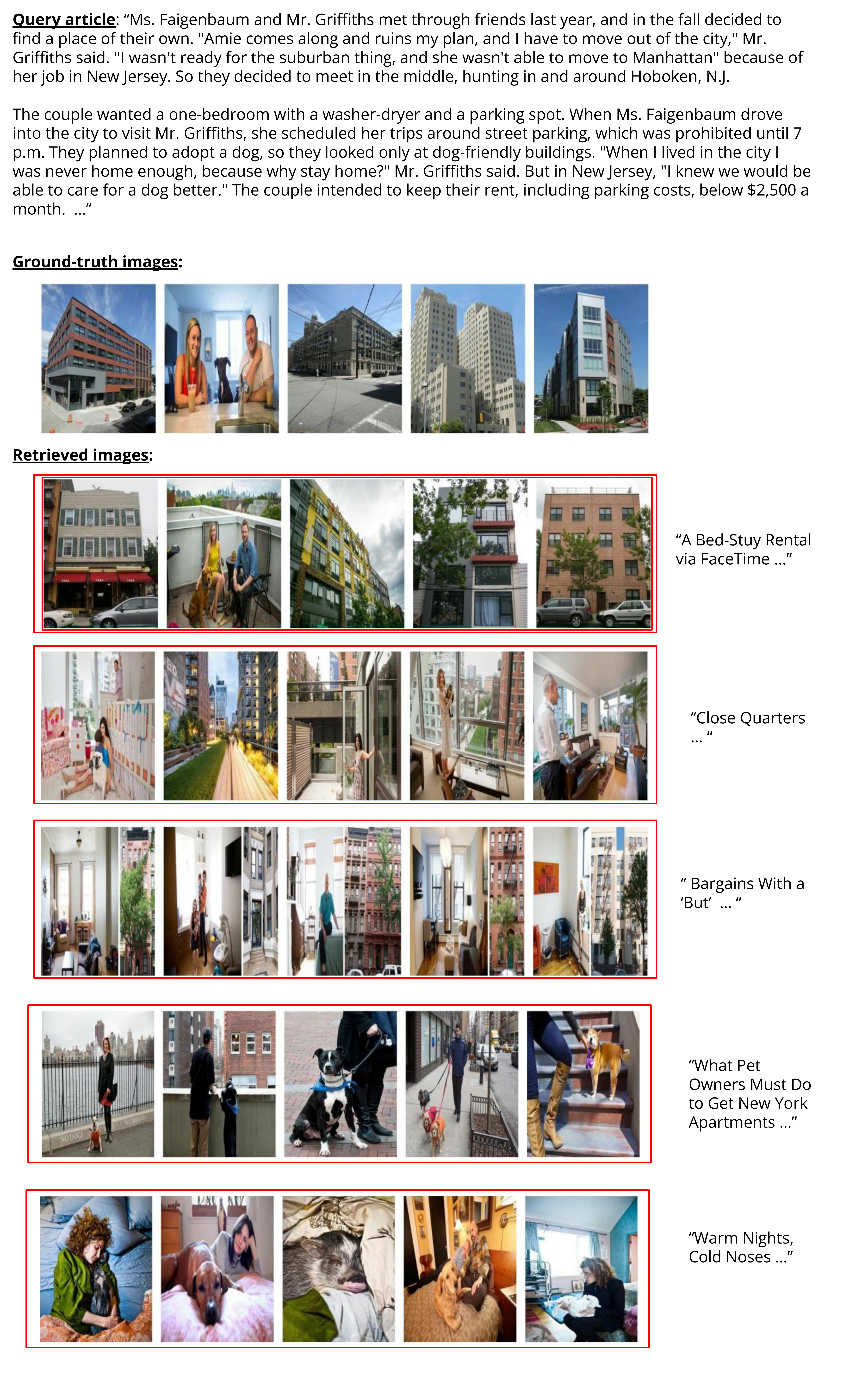}
\vspace{-9cm}
\end{center}
   \caption{Incorrect retrievals on the GoodNews dataset.}
\label{fig:supp_goodnews_qualitative}
\vspace{\fullpagefigurepad}
\end{figure*}

\begin{figure*}[t]
\begin{center}
\includegraphics[width=\linewidth, height=18cm]{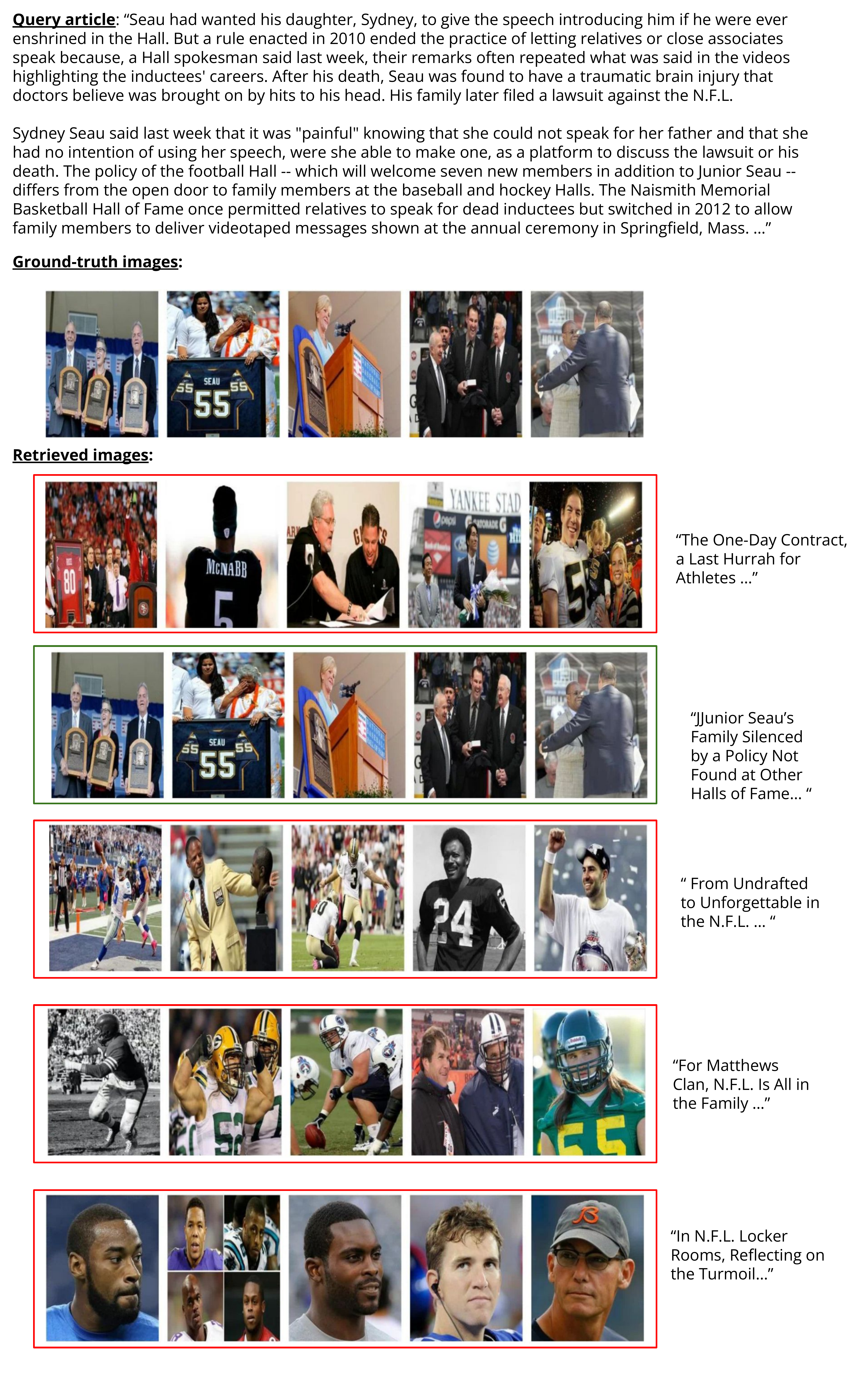}
\vspace{-9cm}
\end{center}
   \caption{Incorrect retrievals on the GoodNews dataset.}
\label{fig:supp_goodnews_qualitative}
\vspace{\fullpagefigurepad}
\end{figure*}

\begin{figure*}[t]
\begin{center}
\includegraphics[width=\linewidth, height=18cm]{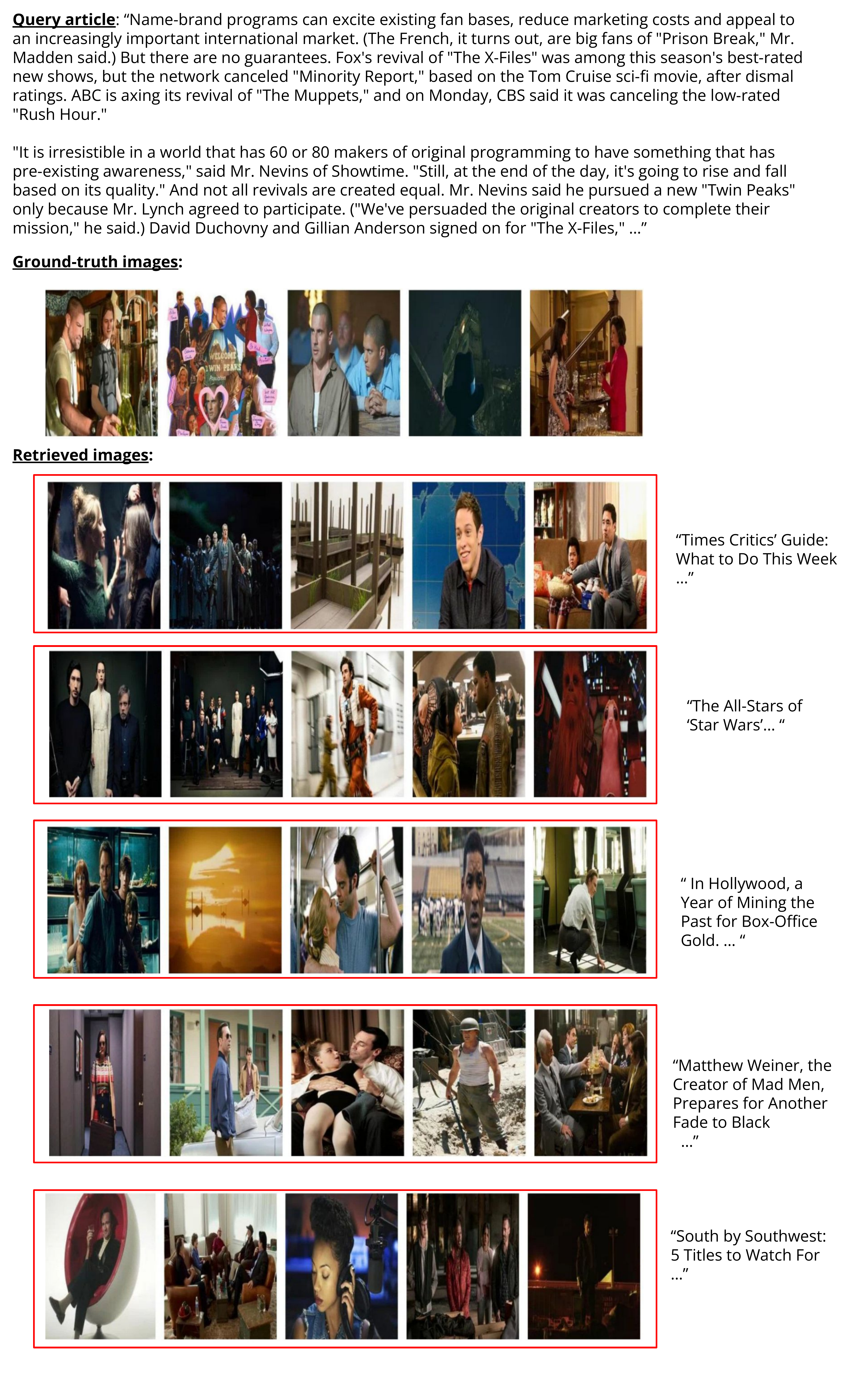}
\vspace{-9cm}
\end{center}
   \caption{Incorrect retrievals on the GoodNews dataset.}
\label{fig:supp_goodnews_qualitative}
\vspace{\fullpagefigurepad}
\end{figure*}

\begin{figure*}[t]
\begin{center}
\includegraphics[width=\linewidth, height=18cm]{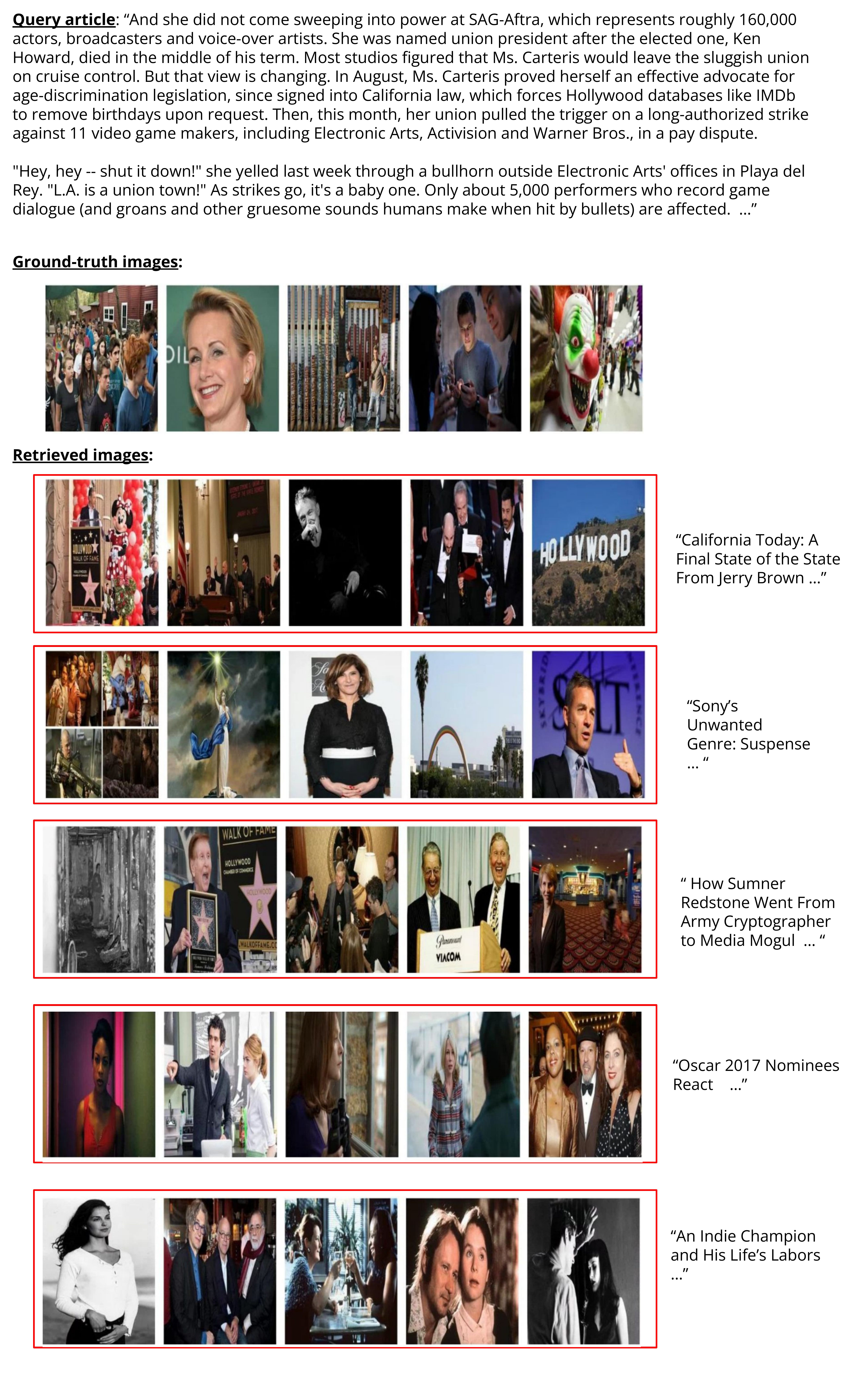}
\vspace{-9cm}
\end{center}
   \caption{Incorrect retrievals on the GoodNews dataset.}
\label{fig:supp_goodnews_qualitative}
\vspace{\fullpagefigurepad}
\end{figure*}

\begin{figure*}[t]
\begin{center}
\includegraphics[width=\linewidth, height=18cm]{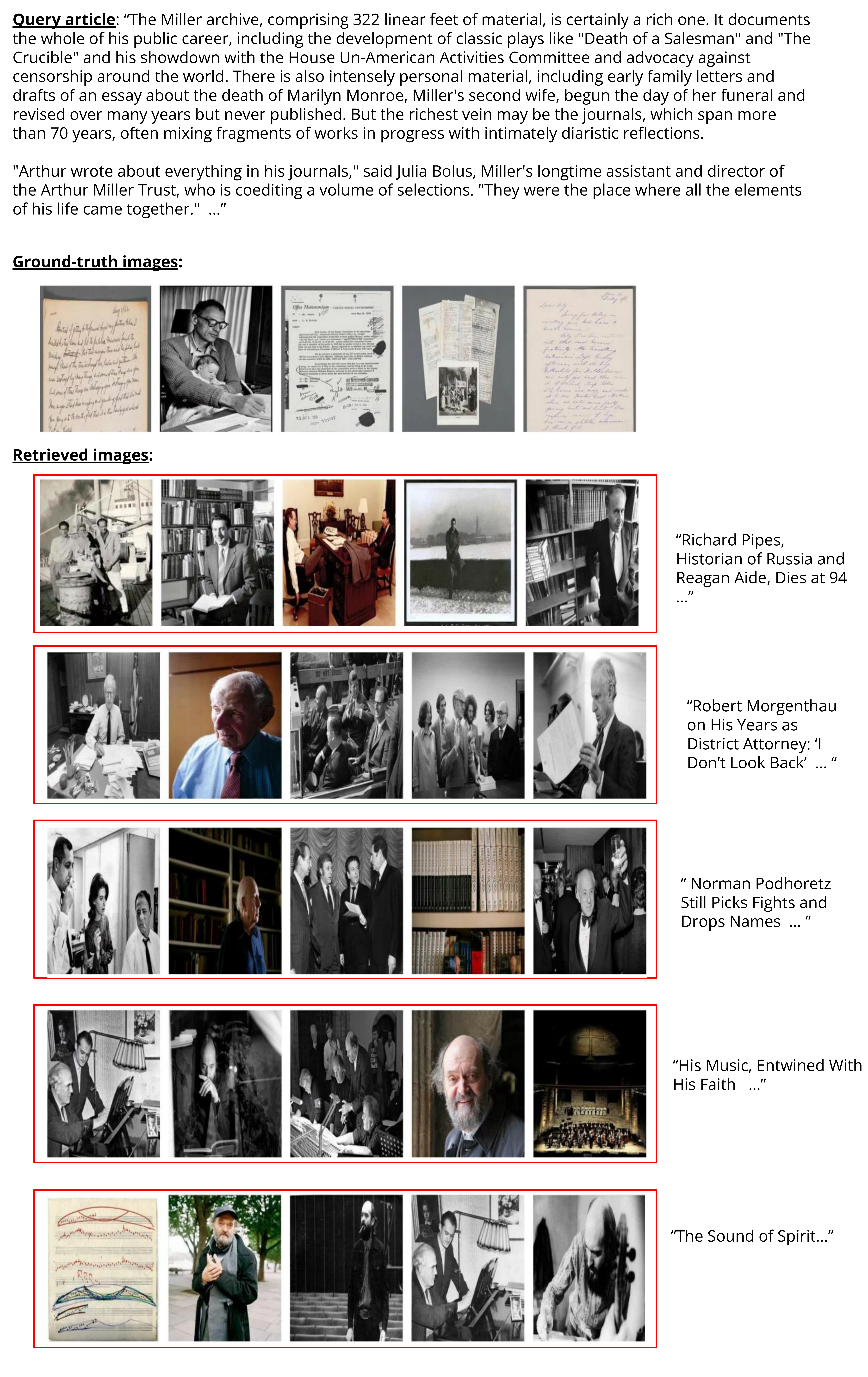}
\vspace{-9cm}
\end{center}
   \caption{Incorrect retrievals on the GoodNews dataset.}
\label{fig:supp_goodnews_qualitative}
\vspace{\fullpagefigurepad}
\end{figure*}

\end{document}